\documentclass{ieeeaccess}
\usepackage{cite}
\usepackage{amsmath,amssymb,amsfonts}
\usepackage{listings}
\usepackage{algorithm}
\usepackage{algpseudocode}
\usepackage{graphicx}
\usepackage{subfig}
\usepackage{url}
\usepackage{newfloat}
\usepackage{multicol}
\usepackage{textcomp}
\usepackage{multirow}
\usepackage{latexsym}
\usepackage{adjustbox}
\usepackage{enumitem}
\usepackage{newfloat}
\usepackage{listings}
\usepackage{makecell}
\usepackage{xspace}
\usepackage{verbatim} 
\usepackage{nicefrac} 

\usepackage{xcolor}
\usepackage{pict2e}

\newsavebox{\ORCIDlogo}
\savebox{\ORCIDlogo}{%
\setlength{\unitlength}{\dimexpr 1em/256\relax}%
\begin{picture}(256,256)%
  \color[HTML]{A6CE39}\put(128,128){\circle*{256}}%
  \color{white}%
  \put(78.6,199.2){\circle*{20}}%
  \moveto(70.9,176,9)\lineto(86.3,176,9)\lineto(86.3,69.8)\lineto(70.9,69.8)%
  \closepath\fillpath%
  \moveto(108.9,176.9)\lineto(150.5,176.9)%
  \curveto(190.1,176.9)(207.5,148.6)(207.5 ,123.3)%
  \curveto(207.5,95,8)(186,69.7)(150.7,69.7)%
  \lineto(108.9,69.7)%
  \closepath\fillpath%
  \color[HTML]{A6CE39}%
  \moveto(124.3,83.6)\lineto(148.8,83.6)%
  \curveto(183.7,83.6)(191.7,110.1)(191.7,123.3)%
  \curveto(191.7,144.8)(178,163)(148,163)%
  \lineto(124.3,163)%
  \closepath\fillpath%
\end{picture}%
}

\newcommand{\gbar}{\overline{\mathbf{g}}}
\newcommand{\gtilde}{\tilde{\mathbf{g}}}

\newcommand{\VM}{\mathcal{V}}

\newcommand{\Alg}[1]{Algorithm~\ref{#1}}
\newcommand{\Eqn}[1]{Eqn (\ref{#1})}

\newcommand{\Thm}[1]	{Thm \ref{#1}}

\newcommand{\dpriv}{\textrm{metric DP}\xspace}

\newcommand{\DirDP}{\textsc{DirDP-SGD}\xspace}

\newtheorem{definition}{Definition}
\newtheorem{theorem}{Theorem}
\newtheorem{corollary}{Corollary}
\newenvironment{proof}{\paragraph{Proof:}}{\hfill$\square$}

\newcommand\orcidicon[1]{\href{https://orcid.org/#1}{\usebox{\ORCIDlogo}}}

\usepackage{hyperref} 

\usepackage{bm}
\makeatletter
\AtBeginDocument{\DeclareMathVersion{bold}
\SetSymbolFont{operators}{bold}{T1}{times}{b}{n}
\SetSymbolFont{NewLetters}{bold}{T1}{times}{b}{it}
\SetMathAlphabet{\mathrm}{bold}{T1}{times}{b}{n}
\SetMathAlphabet{\mathit}{bold}{T1}{times}{b}{it}
\SetMathAlphabet{\mathbf}{bold}{T1}{times}{b}{n}
\SetMathAlphabet{\mathtt}{bold}{OT1}{pcr}{b}{n}
\SetSymbolFont{symbols}{bold}{OMS}{cmsy}{b}{n}
\renewcommand\boldmath{\@nomath\boldmath\mathversion{bold}}}
\makeatother

\def\BibTeX{{\rm B\kern-.05em{\sc i\kern-.025em b}\kern-.08em
    T\kern-.1667em\lower.7ex\hbox{E}\kern-.125emX}}

\begin{document}
\history{Date of publication xxxx 00, 0000, date of current version xxxx 00, 0000.}
\doi{10.1109/ACCESS.2024.0429000}

\title{Empirical Calibration and Metric Differential Privacy in Language Models}
\author{\uppercase{Pedro Faustini} \orcidicon{0000-0002-9961-1252}\authorrefmark{1},
\uppercase{Natasha Fernandes}\orcidicon{0000-0002-9212-7839}\authorrefmark{1}, 
\uppercase{Annabelle McIver}\orcidicon{0000-0002-2405-9838}\authorrefmark{1},\uppercase{Mark Dras}\orcidicon{0000-0001-9908-7182}\authorrefmark{1},}

\address[1]{Macquarie University, Sydney, NSW 2109 Australia}
\tfootnote{\textbf{This work has been submitted to the IEEE for possible publication. Copyright may be transferred without notice, after which this version may no longer be accessible.}}

\markboth
{Faustini \headeretal: Empirical Calibration and Metric Differential Privacy in Language Models}
{Faustini \headeretal: Empirical Calibration and Metric Differential Privacy in Language Models}

\corresp{Corresponding author: Pedro Faustini (e-mail: pedro.faustini@mq.edu.au).}

\begin{abstract}

  NLP models trained with differential privacy (DP) usually adopt the DP-SGD framework, and privacy guarantees are often reported in terms of the privacy budget $\epsilon$. 
However, $\epsilon$ does not have any intrinsic meaning, 
and it is generally not possible to compare across variants of the framework.  Work in image processing has therefore explored how to \textit{empirically calibrate} noise across frameworks using Membership Inference Attacks (MIAs).  However, this kind of calibration has not been established for NLP.  In this paper, we show that MIAs offer little help in calibrating privacy, whereas reconstruction attacks are more useful.
As a use case, we define a novel kind of \textit{directional privacy} based on the von Mises-Fisher (VMF) distribution, a metric DP mechanism that perturbs angular distance rather than adding (isotropic) Gaussian noise, and apply this to NLP architectures. We show that, even though formal guarantees are incomparable, empirical privacy calibration reveals that each mechanism has different areas of strength with respect to utility-privacy trade-offs.
\end{abstract}


\titlepgskip=-21pt

\maketitle

\section{Introduction}\label{sec:introduction}




Research has shown how Differential Privacy (DP) \cite{dwork-roth:2014}, and DP-SGD specifically as a method for training machine learners \cite{abadi-etal:2016:CCS}, can protect sensitive information, including in NLP tasks \cite{hoory-etal-2021-learning,kerrigan-etal-2020-differentially}. Works usually adopt the Laplace or Gaussian mechanisms, where noise determined by the mechanism type and a privacy budget $\epsilon$ is added to the output of the model, to its input data, or to the gradients during training. The challenge is finding an acceptable tradeoff between preserving privacy and keeping utility.



Various work has observed \cite{dwork-etal:2019} that it is difficult to know what a privacy budget $\epsilon$ `means', and how to choose one.  \cite{DBLP:conf/uss/Jayaraman019} consequently proposed a \textit{calibration} of DP with respect to a concrete empirical task that captures privacy leakage; this additionally allows clearer comparison and understanding of privacy-utility trade-offs 
where various DP frameworks (e.g., R\'enyi DP \cite{8049725}) may be used in privacy accounting.\footnote{Note that this distinct from the goal of privacy auditing: the early work of \cite{jagielski2020auditing} characterises auditing as aiming to determine precise privacy guarantees of DP-SGD, and contrasts it with the goal of \cite{DBLP:conf/uss/Jayaraman019} of comparing privacy guarantees across frameworks.}  Their proposal to use Membership Inference Attacks (MIAs), where success in attacking a machine learning model indicates potential privacy leakage in the form of knowing whether an individual is in the training set or not, has been widely taken up in the image processing literature \cite{8962136, pmlr-v139-choquette-choo21a, watson2022on}. 

\begin{figure}
        \centering
        \includegraphics[scale=0.6]{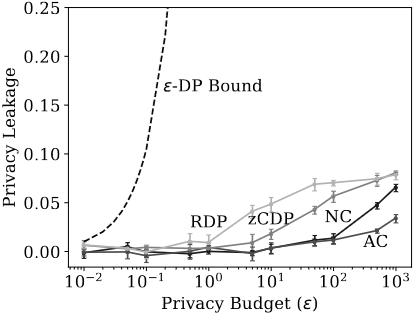}
        \caption{\cite{DBLP:conf/uss/Jayaraman019}'s Fig 2(a): calibration plotting $\epsilon$ against privacy leakage for several DP-SGD variants.}
        \label{fig:jayaraman_example}
    \end{figure}

Usually, works presuppose that the attack is `well-behaved' with respect to $\epsilon$: smaller privacy budgets, and consequent larger noise, should correspond roughly monotonically to less effective attacks, so that effects of $\epsilon$ values can be distinguished and compared within and across DP variants. The image data of \cite{DBLP:conf/uss/Jayaraman019} and subsequent work follows this pattern (see Fig~\ref{fig:jayaraman_example}). However, a calibration in the style of \cite{DBLP:conf/uss/Jayaraman019} has not been established for language data, despite MIAs having been used as supplementary evaluations for empirically assessing privacy leakage in NLP works (for example, \cite{krishna-etal-2021-adept,vakili-dalianis-2023-using}). Suggesting that a straightforward use as in the image domain is unlikely, some other work \cite{senge-etal-2022-one} has shown that DP may have its peculiarities when applied to language data, and further, some very recent work has suggested that MIAs in general work poorly against large language models \cite{duan2024membershipinferenceattackswork}.

We note that membership inference is not the only attack type suitable for calibration; another potential type of attack is the reconstruction of training data, with \cite{dwork-etal:2017} characterising it as particularly important.  In this paper, then, we investigate the behaviour of both MIAs and reconstruction attacks with respect to DP-SGD on NLP datasets, to see how useful they are for calibration in these contexts. 

As noted, an empirical calibration allows the comparison of noise mechanisms outside the standard DP-SGD framework. One example is metric DP (which generalises the distance function; standard DP is then the special case of metric DP with the Hamming distance), which has been seen in NLP works mostly dealing with local DP (for instance, \cite{fernandes-etal:2019:POST, feyisetan-etal:2019, SilvaCarvalho2023}).
Gaussian noise is isotropic: its noise vector is equally likely to point in any direction in the high-dimensional space of the gradients. In contrast, we might expect that the utility of the model would be better served by a mechanism which is designed to preserve the \emph{direction} of the gradients. Intuitively, the more their directions are preserved, the better is the gradient descent algorithm going to minimise the loss function.

The idea of tuning the \emph{shape} of the noise arises from \emph{metric differential privacy} or $d$-privacy~\cite{chatzikokolakis-etal:2013:PETS}, a generalisation of DP in which the notion of adjacency is generalised to a distinguishability metric $d$. Metric differential privacy is a unifying definition which subsumes both central and local DP, the former recoverable by choosing $d$ to be the Hamming metric on databases, and the latter by choosing $d$ to be the Discrete metric on individual data points. By careful choice of the metric $d$, $d$-privacy mechanisms can provide a better privacy-utility trade-off than with standard DP. We thus compare the Gaussian-noise mechanism of \cite{abadi-etal:2016:CCS} with the (directional privacy) von Mises-Fisher mechanism of  \cite{weggenmann-kerschbaum:2021:CCS}, who applied it to recurrent temporal data rather than deep learning.

As a use case of our calibration framework, then, we define and apply for the first time a directional privacy mechanism to fine-tuning Transformers and to NLP tasks, and compare it to the standard DP-SGD.


The paper is structured as followed.  In Sec~\ref{sec:related_work}, we review related work on use of DP in ML and specifically NLP, and also different kinds of model attacks and how privacy has been calibratd.  In Sec~\ref{sec:calibration}, we assess the suitability of different attacks for privacy calibration in NLP.  In Sec~\ref{sec:privacy_model}, we outline a novel application of metric DP for privately training NLP models; we then empirically compare this metric DP approach to standard baselines using the privacy calibration established in Sec~\ref{sec:calibration}.  We conclude in Sec~\ref{sec:conclusions}.

In this paper, our contributions in summary are as follows.

\begin{enumerate}
    \item We show that MIAs are not well suited for calibration of DP in language data.  We also show that reconstruction attacks provide a well-behaved (albeit limited) alternative.
    
    \item We define and apply a metric DP mechanism based on angular distance (via the von Mises-Fisher distribution) for the first time in training Transformer models.  A comprehensive set of experiments assesses how both this new mechanism and standard DP-SGD perform in preventing such attacks in NLP. We show that VMF outperforms Gaussian in some contexts.
\end{enumerate}
\section{Related work}\label{sec:related_work}

In this section, we review relevant work on the use of DP in deep learning (\S\ref{sec:lit-rev-DP-deep}), and on \dpriv, including its own intersections with deep learning.

\subsection{Differential Privacy in Deep Learning}
\label{sec:lit-rev-DP-deep}
    Neural networks can be victims of several types of attacks, like membership inference \cite{Shokri2016MembershipIA, 8962136}, model stealing \cite{Yu2020CloudLeakLD} and data reconstruction \cite{zhu-etal:2019:NeurIPS, DBLP:journals/corr/abs-2001-02610, geiping-etal:2020:NeurIPS, wei-etal:2020:ESORICS}. This motivates the need for privacy guarantees to protect neural networks while keeping their utility for the task they are trained to deal with.
    
    \cite{song-etal:2013} proposed  Differentially Private Stochastic Gradient Descent (DP-SGD), which first brought DP to the training of gradient-descent models. DP-SGD adds calibrated noise in the gradients during training, before updating the parameters. 
    This was followed by works that looked at providing efficient algorithms and tightening error bounds \cite[for example]{bassily-etal:2014:FOCS} so that the addition of noise would not degrade utility to impractical levels.  A key work in this direction was made by \cite{abadi-etal:2016:CCS}, who introduced a technique to keep track of the privacy budget, called the Moments Accountant, specifically for the Gaussian mechanism.
    
    The additional privacy comes with a cost, in that the noisy gradients may affect the utility of the model. Therefore, either better features may be collected or handcrafted, or even more data may be needed \cite{tramer2021differentially}. \cite{li-etal:2022:ICLR} (in NLP) and \cite{de-etal:2022} (in computer vision) also found that DP-SGD can perform well in very different regions of the hyperparameter space relative to non-private models.  The architecture of the model may also play a role in the utility, with larger and pretrained models being more efficiently fine-tuned, especially with larger batch sizes \cite{li-etal:2022:ICLR, DBLP:journals/corr/abs-2108-01624}, which can be computationally demanding.

    Proposals to change the DP-SGD algorithm itself have also been made, many relating to clipping strategies. \cite{10.1145/3447548.3467268} observed that clipping and noise addition affect underrepresented classes, making the accuracy of the model for them even lower. Thus they proposed to control the contribution of samples in a group according to the group clipping bias. \cite{10.1145/3469877.3490594} proposed to divide gradients from $m$ samples into $k$ groups. The gradients in each group are clipped with a different bound, as opposed to a global bound from DP-SGD. They argue that clipping could distort gradient information. 
    
    However, all these works have adopted isotropic noise, often from the Gaussian distribution. Clipping the gradients derived from these noises limits their \emph{length}, but does not alter their \emph{direction}. There is a lack of studies comparing how different noise distributions affect the privacy/utility trade-off and how noise distributions other than isotropic ones can be used during the training of neural networks.

\subsubsection{Differential Privacy in NLP}
    
   In NLP, \cite{kerrigan-etal-2020-differentially} fine-tuned GPT2 \cite{radford2019language} and two other feed-forward neural networks with the Gaussian mechanism, and measured the perplexity of the generated text. \cite{hoory-etal-2021-learning} trained (from scratch) BERT \cite{devlin-etal-2019-bert} with a modified and differentially private word-piece algorithm, to create a new domain-specific vocabulary (in their case, from clinical data). They further fine-tuned the model with DP-SGD and concluded that their DP-BERT prevented memorisation of sequences of words.  \cite{klymenko-etal:2022:differential} and \cite{hu-etal-2024-differentially} provide surveys of work since then.
    
    \cite{senge-etal-2022-one} analysed the behaviour of BERT and LSTM in several downstream tasks. They found unexpected phenomena, like low $\epsilon$ making the model biased towards predicting the same class for sentiment analysis and that the performance drop with decreasing $\epsilon$ is not a consistent pattern, sometimes barely happening at all.

\subsection{Attacks against privacy}

    Originally applied to statistical databases, DP states that an analysis reaches the same conclusions regardless if any \textit{individual} is removed from the dataset \cite{dwork-roth:2014}. Thus, the output of a mechanism $\mathcal{M}$ should be very similar if it received a dataset $x$ or \emph{neighbouring} one $x'$ (i.e., $x, x' \in \mathcal{X}$ differ in one element). This is done by adding noise calibrated by a parameter $\epsilon{ >} 0$. The smaller the $\epsilon$, the more noise is added.

    Assessing how protected a model is in practice often needs some empirical assessment. 
    \cite{DBLP:conf/uss/Jayaraman019} therefore used attacks against a model as a way of understanding (a lower bound on) privacy leakage. They argued that in iterated learning contexts, various relaxed definitions of DP --- Concentrated DP \cite{dwork-rothblum:2016}, Zero Concentrated DP \cite{bun-steinke:2016} and R\'enyi DP \cite{mironov:2017:CSF}, as compared to basic and advanced composition --- present poorly understood trade-offs between privacy and utility, and so proposed empirical attacks as a way of calibrating across DP variants. They studied Membership Inference Attacks and, to a lesser extent, attribute inference or model inversion attacks.  They measured the amount of information leakage via attacker advantage as defined in \cite{DBLP:conf/csfw/YeomGFJ18}, 
        experimentally deriving these scores for the image dataset CIFAR100 \cite{Krizhevsky2009LearningML} and the tabular dataset Purchase 100 \cite{acquire-valued-shoppers-challenge}.  
        This allowed a comparison across the DP variants, as in Fig~\ref{fig:jayaraman_example} taken from their paper: it can be seen that R\'enyi DP leaks the most for a given $\epsilon$ (most noticeable at higher $\epsilon$).
        They concluded that relaxed definitions of differential
privacy that reduce the amount of noise needed to improve
utility also increase the measured privacy leakage.
In drawing their conclusion, they are able to carry out this calibration because they observed that as expected, ``the differential privacy variations have leakage in accordance with the amount of noise they add for
a given $\epsilon$,'' which is what we referred to as `well-behaved' above.  We investigate in this paper whether this is the case in NLP datasets.

    \textbf{Membership inference attacks} \cite{Shokri2016MembershipIA} infer whether an object was used for training by looking at the class probability outputs of a target model. This attack has several versions \cite{hu-etal:2022}. Usually, numerous \emph{shadow models} try to mimic the target model, and their outputs are used to train a classifier that judges whether a data point was in the training set or not.  \cite{10.1145/3548606.3560675} proposed an enhancement (\textbf{MIA-R}) which we use in our experiments where the loss of the target model is compared against a threshold function built after training several reference (akin to shadow) models. Usually, overfitted models are better exploited by this attack \cite{DBLP:conf/csfw/YeomGFJ18, 8962136}. 
    \cite{DBLP:conf/uss/Jayaraman019} used the attacks by \cite{Shokri2016MembershipIA} and \cite{DBLP:conf/csfw/YeomGFJ18} in their work.

    Recently, in the NLP space, \cite{mattern-etal-2023-membership} proposed a method that does not require training reference models. Instead, they artificially generate synthetic neighbouring samples and compare the difference model's loss against the true sample and the average of the synthetic ones. In our experiments, we refer to it as \textbf{MIA-N}.
    Working with medical data, and looking at the kinds of samples that are ``memorised,'' \cite{mireshghallah-etal-2022-quantifying} proposed a new likelihood-based MIA for ClinicalBERT. There has also been debate about how suitable these attacks are to quantify privacy risks for some tasks. For instance, \cite{vakili-dalianis-2023-using} deployed MIA against models trained on a clinical dataset that had all its names replaced, and their work focuses on protecting the names. The target model was trained with sensitive data, but the reference model was not. Their results suggested that MIA does not accurately measure privacy gains from using pseudonymisation: MIA may quantify privacy at datapoint level, 
    but not token level.  

    There has been recent interest in developing bounds for DP w.r.t. data reconstruction \cite{guo-etal:2022:PMLR,hayes-etal:2024:NeurIPS}, although again so far only with image data like CIFAR-10.  We therefore consider here \textbf{gradient-based reconstruction} of data, as an alternative to MIAs.  They work in a setting where several parties work together to train a model, and for privacy reasons the datasets are not shared; instead, the gradients are exchanged. \cite{zhu-etal:2019:NeurIPS} showed that it is possible to reconstruct images by minimising the distance between gradients generated from random inputs and labels to the received gradients. This family of attacks was enhanced \cite{geiping-etal:2020:NeurIPS, NEURIPS2021_0d924f0e} and then expanded to reconstruct text \cite{NEURIPS2022_35b5c175, deng-etal-2021-tag-gradient, fowl2022robbing, NEURIPS2022_32375260}. DP has been shown to be able to provably thwart such attacks, as in \cite{dwork-etal:2017}. 
    
    In NLP, versions exploiting specific architectures were also proposed, e.g., the Decepticons \cite{fowl2023decepticons} and LAMP \cite{NEURIPS2022_32375260} attacks, which target Transformers and BERT respectively; we adopt these in our work given they are relatively recent and widely used, and have solid results.

\section{Privacy Calibration}
\label{sec:calibration}

This section assesses the suitability of MIAs versus reconstruction attacks for calibrating privacy in NLP.  To do this, we train classification models across different datasets under various levels of privacy. Following \cite{DBLP:conf/uss/Jayaraman019}, we compare how models are susceptible to different flavours of MIAs.
Foreshadowing the results, we notice that MIA effectiveness is limited due to reliance on overfitting. Therefore, we resort to gradient-based reconstruction attacks and show they present more meaningful patterns for privacy calibration.

\subsection{Experimental setup}\label{sec:setup}

    \paragraph{Datasets and models} We used GPT2 and BERT uncased (pretrained base versions) models from the Transformers library \cite{wolf-etal-2020-transformers}. We thus follow the choice of other works which also studied DP-SGD in NLP \cite{kerrigan-etal-2020-differentially, hoory-etal-2021-learning, senge-etal-2022-one}. We fine-tune the classifier head and pooler layers of BERT. Thus, we fine-tune 592,130 parameters out of $\sim$108M. For GPT2, we fine-tune the scorer and the normalisation of the final layer. Thus, we fine-tune 3,072 parameters out of $\sim$124M. Freezing most layers and using only the top ones has also been done in previous works \cite{senge-etal-2022-one}.

    For the amount of Gaussian noise, we follow previous works in NLP, by taking noise multipliers that lead to $\epsilon$ ranging from single digits, as in \cite{hoory-etal-2021-learning}, to budgets over one hundred \cite{kerrigan-etal-2020-differentially} or even thousands \cite{igamberdiev2023dpbart}.  We cover all three tiers of protection described by \cite[\S5.2]{ponomareva-etal:2023:JAIR}. We fix the target $\epsilon$ for the utility experiments and report the noise multiplier $\sigma$ returned by the Renyi privacy accountant, using it in the subsequent experiments. Table \ref{tab:budgets} maps $\sigma$ to $\epsilon$. We set the clipping parameter C=1. 

    We used the following datasets: 1) IMDb \cite{maas-etal-2011-learning} and SST2 \cite{socher-etal-2013-recursive}, which are datasets for sentiment classification and 2) CoLA \cite{DBLP:journals/corr/abs-1805-12471}, which contains grammatical and ungrammatical sentences. For the SST2 and IMDb datasets, we split the original training set into 90\%/10\% for training and validation subsets respectively. For the CoLA dataset, which is heavily imbalanced, we undersampled the training and validation sets to make both classes equivalent, but left the test set unchanged. More details about the train/validation/test split can be found in Appendix \ref{app:datasets_splits}.

    \paragraph{Utility experiments}\label{sec:utility_setup}

    We fine-tune BERT and GPT2 on different datasets and compare their performance across classification tasks under different privacy settings. We then use the same noise multiplier for the subsequent experiments.
    
    For IMDb and SST2, we measure the accuracy of the models. Since CoLA has an imbalanced test set, we report the Matthew's correlation coefficient (MCC) \cite{MATTHEWS1975442}. A value of 1 means perfect prediction, 0 is random and -1 indicates an inverse prediction.

    \paragraph{Membership Inference Attack} This attack infers whether a test sample was used during training based on the model's output. There are several variants for this attack. We use the vanilla the Membership Inference Attack (MIA) \cite{Shokri2016MembershipIA} and its extension via Reference Models (MIA-R) from the ML Privacy Meter\footnote{\url{https://github.com/privacytrustlab/ml_privacy_meter}} \cite{nasr-etal:2018:SP,kumar-shokri:2020,10.1145/3548606.3560675} library and trained 10 shadow/reference models. 
    They follow the architecture and hyperparameters of the target models, but are not trained with differential privacy. More details about dataset splits can be found in Appendix \ref{app:datasets_splits}.
    Last, we deploy the MIA-N of \cite{mattern-etal-2023-membership}. This attack has the advantage of not requiring shadow models because the neighbouring samples are artificially generated, so there is no need for splitting the datasets.
    
    We measure the attack success by the Area Under the Curve (AUC) metric, computed by plotting a ROC curve representing the trade-off between False Positive Rate (FPR, or classifying \emph{out} samples as \emph{in} samples) and True Positive Rate (TPR, or correctly identifying \emph{in} samples). The attack is stronger when the AUC is bigger.  \cite{mireshghallah-etal-2022-quantifying} and others use this as a measure of privacy loss.

    To be able to compare more directly with \cite{DBLP:conf/uss/Jayaraman019}, we also measure the vanilla MIA success by computing the privacy leakage metric that they used and that was proposed by \cite{DBLP:conf/csfw/YeomGFJ18}. It is the difference between the TPR and FPR rates of the inference attack, and it ranges from 0 to 1. Small values indicate low levels of leakage.

    \paragraph{Text reconstruction from gradients} This attack recovers original sentences used to train a model when gradients are shared in a Federated Learning environment. There are several variants from this attack, as discussed in Section \ref{sec:related_work}. The \textbf{Decepticons} attack assumes an untrusted server that sends the state of a malicious model and then recovers user data from their updates.
    
    Decepticons\footnote{\url{https://github.com/JonasGeiping/breaching}} exploits characteristics of the Transformer architecture and the token embeddings: the attacker disables the attention layers and the outputs of each feed-forward block (except the last entry). It aims to reconstruct the inputs to the first transformer layer, (the token embeddings and positional embeddings), which are also the largest layers. We thus add DP noise to these large layers.
    
    The \textbf{LAMP} attack\footnote{\url{https://github.com/eth-sri/lamp}} \cite{NEURIPS2022_32375260} follows another family of attacks which uses an auxiliary language model for modelling prior text probabilities and alternates continuous optimisation steps with discrete transformations over the text under reconstruction.


    We assess the effectiveness of the reconstruction with the following measurements:

    \begin{itemize}
        \item \textbf{Jaccard similarity} \cite{f6902328-1c27-32dc-a3cd-6b0b3a4279b1} measures the similarity between two strings by comparing their intersection over their union. It is defined as:

        $$Jaccard(A, B) = \frac{A \cap B}{A \cup B}$$

        where  A and B are sets of words from two texts. Higher values indicate greater similarity. Jaccard similarity is useful for evaluating word-level overlap but does not consider semantic meaning. It has been used in other reconstruction works, such as \cite{285479, 10.1145/3564625.3564628}.

    \item \textbf{Cosine similarity} measures the cosine of the angle between two vectors in a high-dimensional space. It is defined as 

    $$CosSim(A, B) = \frac{AB}{||A|| ||B||}$$

    We apply it to the Universal Sentence Encoder (USE) \cite{cer-etal-2018-universal} to capture semantic similarity over two sentence embeddings A and B. It was adopted for evaluating reconstructions by \cite{DBLP:journals/popets/GuKRVM23, chu-etal-2024-reconstruct}.

    \item \textbf{Meteor} \cite{lavie-agarwal-2007-meteor} measures the similarity between a reformulated text and a reference. It aligns words between the two texts using exact matches, stemmed matches, and synonyms before computing a score that considers precision, recall, and a penalty for incorrect word order as:

    $$Meteor = \frac{10 PR}{R + 9P} (1 - penalty)$$

    where P and R are precision and recall based on 1-gram matches and 
penalty penalises word order differences. Meteor was used by previous works in text reconstruction as \cite{chu-etal-2024-reconstruct}.

    \item \textbf{Rouge-L} \cite{lin-2004-rouge} evaluates the quality of a generated text by measuring the longest common subsequence between the generated text and the reference text. It accounts for sequence order without requiring strict contiguity. It is a recall-oriented metric since it emphasises how much of the reference  is captured by the reformulated text. It is defined as:

    $$RougeL(A, B)= \frac{LCS(A, B)}{max(||A||, ||B||)}$$

    where LCS stands for longest common subsequence. It was the standard metric that the authors of LAMP adopted for evaluating their attack, as well in many other reconstruction works \cite{chu-etal-2024-reconstruct, li-etal-2024-seeing}.
        
    \end{itemize}


\subsection{Results}

    \begin{figure*}[ht!]
        \centering
        \small
        \begin{tabular}{ccc}
                   
            \subfloat[GPT2 on CoLA]{\includegraphics[width=0.3\textwidth]{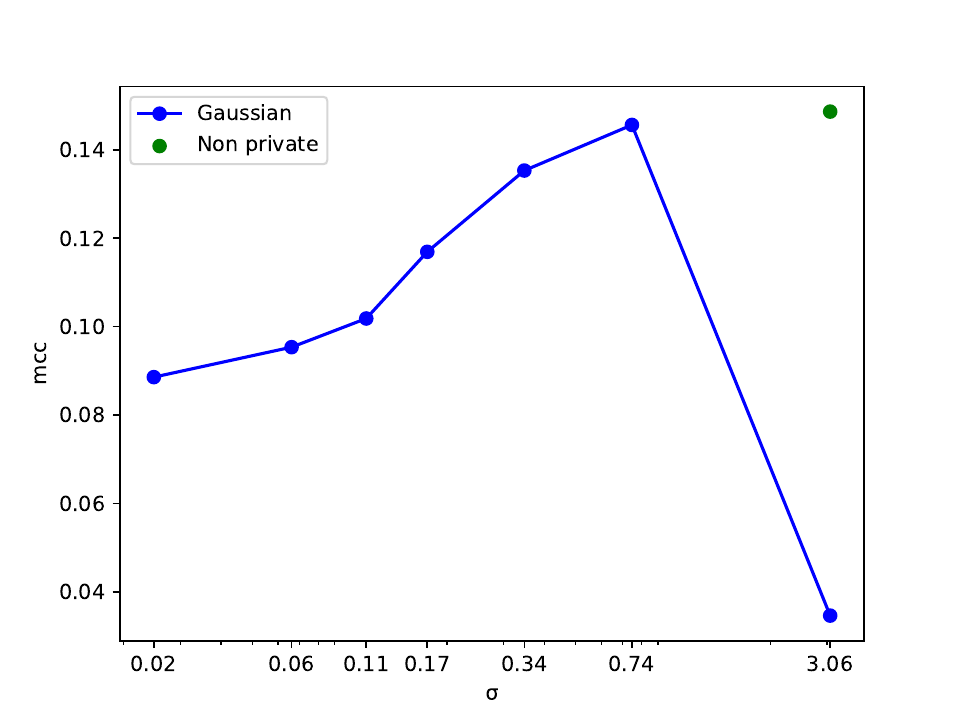}} &
            
            \subfloat[GPT2 IMDb]{\includegraphics[width=0.3\textwidth]{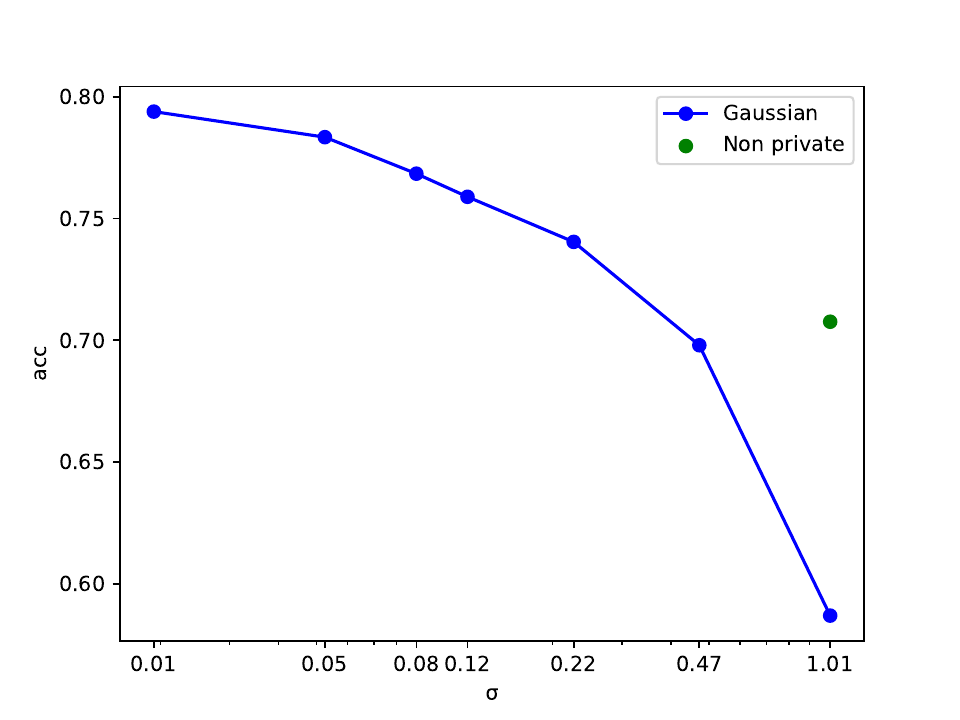}} &
        
            \subfloat[GPT2 SST]{\includegraphics[width=0.3\textwidth]{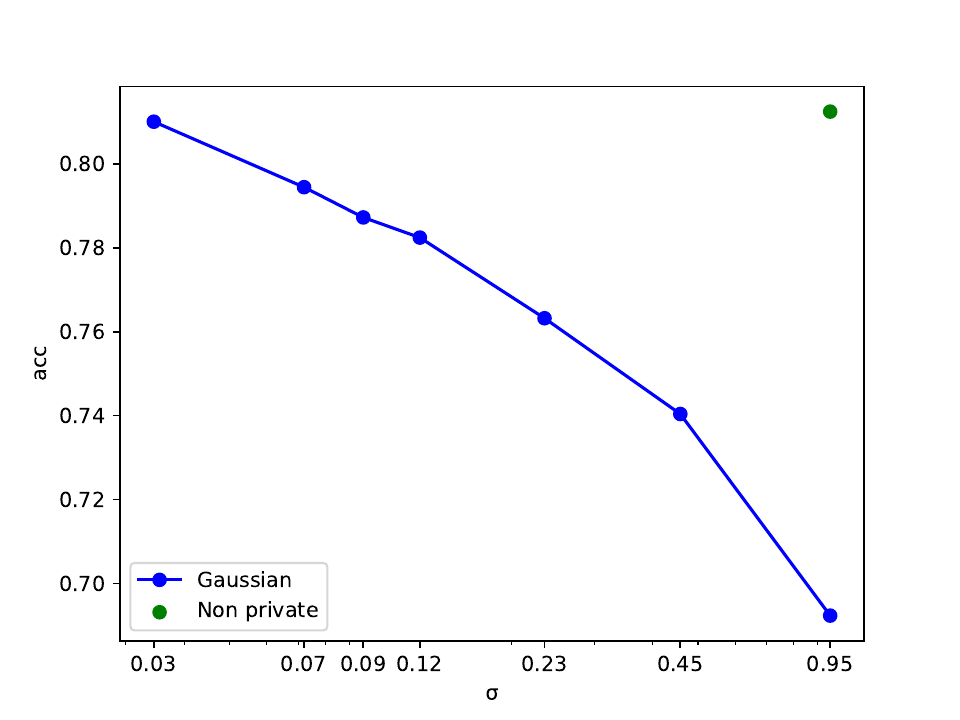}}  \\

            \subfloat[BERT CoLA]{\includegraphics[width=0.3\textwidth]{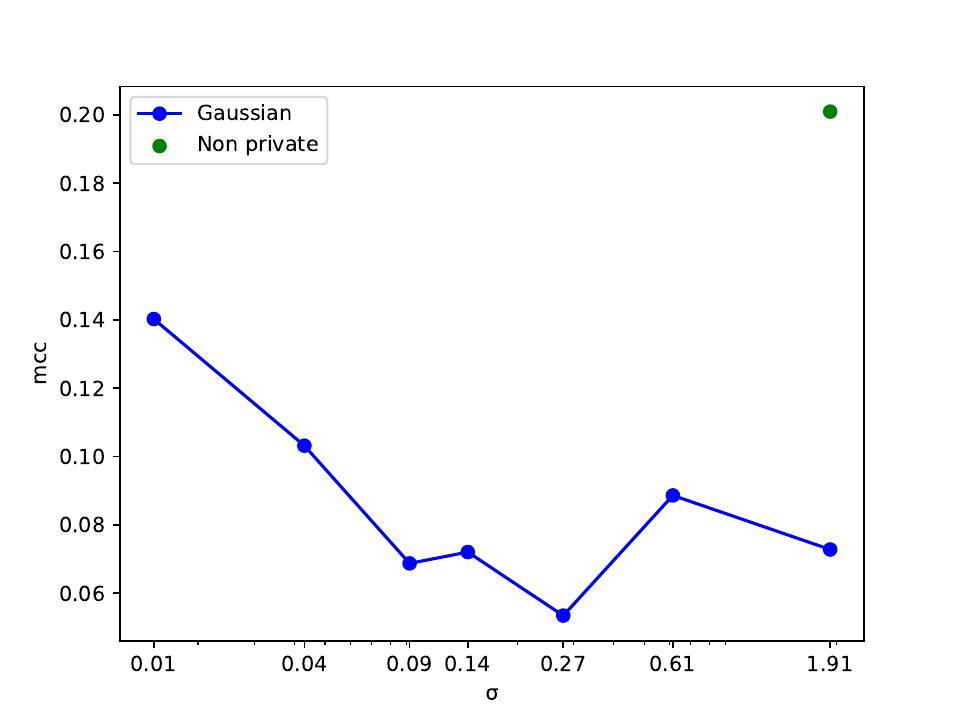}} &
            
            \subfloat[BERT IMDb]{\includegraphics[width=0.3\textwidth]{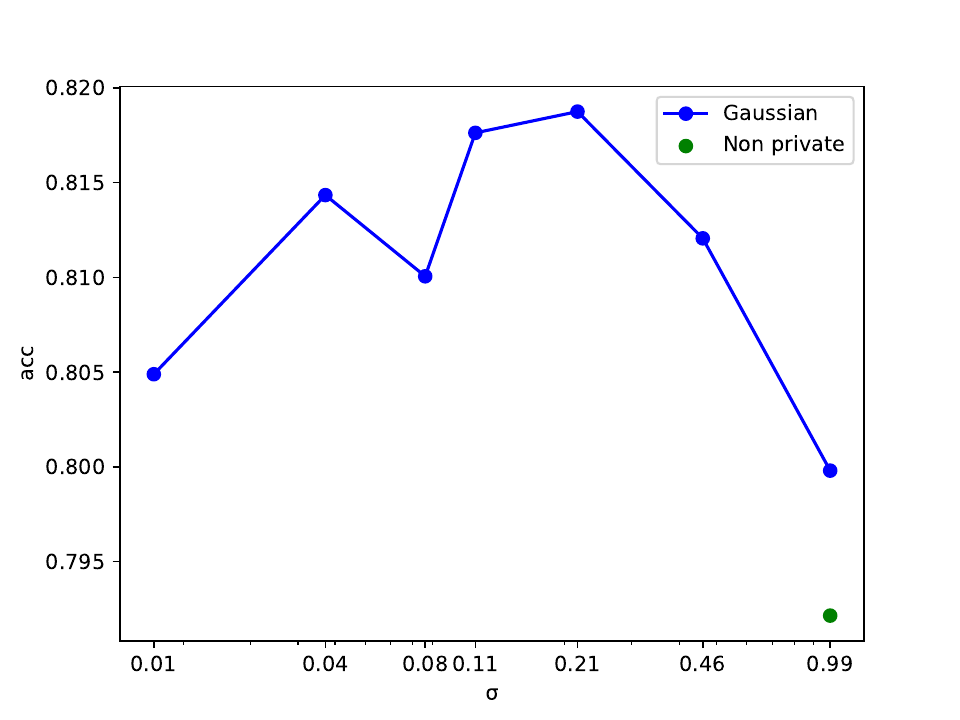}} &
            
            \subfloat[BERT SST]{\includegraphics[width=0.3\textwidth]{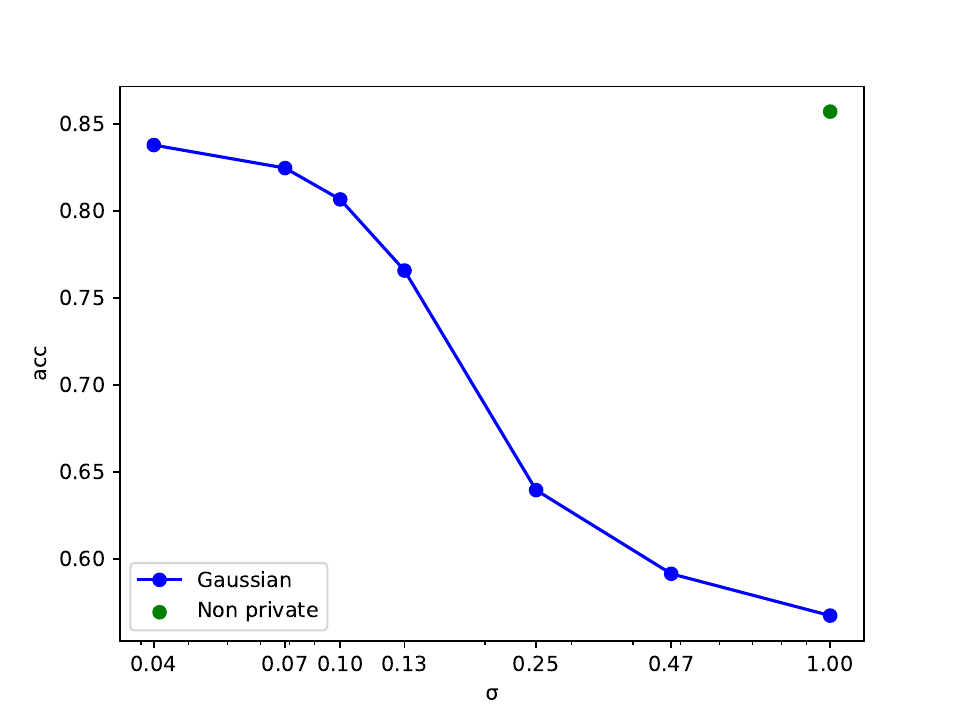}}

        \end{tabular}
        \caption{\textmd{Utility (accuracy or MCC) for models under different privacy settings across each dataset.}}\label{fig:utility_gauss}
        \end{figure*}

    \paragraph{Utility}
    We first report the performance of the models across classification tasks in different datasets, in terms of accuracy (for IMDb and SST2) or MCC (for CoLA) in Figure \ref{fig:utility_gauss}.

    There is a general trend applying to both models on SST2 that increasing noise leads to a decreased accuracy, although the relationship is not perfectly monotonic (noting that this is in line with the results of \cite{senge-etal-2022-one}).
    There is no free lunch for either model, and tighter privacy guarantees yield lower accuracy, going from above 80\% to smaller than 70\%. 
    
    The same does not happen so linearly in CoLA, which is heavily imbalanced dataset. Both models attain only a modest MCC, and their behaviour when noise is added becomes erractic. For IMDb, BERT's accuracy fluctuates around 80\%, whilst for GPT2 it decreases more sharply as the noise increases.
    Finally, we notice that the accuracy for IMDb is worse than many private models; we hypothesised that the DP noise may act mostly as a regularisation factor \cite{li-liu:2020,li-liu:2021}.%
    \footnote{We carried out a small experiment with dropout to verify that this could be the case; we omit for reasons of space.}

   \begin{table*}[h!]
       \centering
       \small
       \begin{tabular}{ccccc|cc|cccc|cc}
       \hline
       &\multicolumn{6}{c|}{\textbf{GPT2}}&\multicolumn{6}{c}{\textbf{BERT}} \\
       
       &&\multicolumn{3}{c|}{\textbf{MIA}}&\multicolumn{2}{c|}{\textbf{MIA-R}} &
       \multicolumn{4}{c|}{\textbf{MIA}}&\multicolumn{2}{c}{\textbf{MIA-R}}\\
       
       \textbf{Dataset} &
       $\textbf{\sigma }$ ($\textbf{\epsilon}$) &
       
       \textbf{AUC}  &
       \makecell{\textbf{Privacy} \\ \textbf{Leakage}}
       & \makecell{\textbf{Train} \\ \textbf{gap}} &
       \textbf{AUC}  &
       \makecell{\textbf{Train} \\ \textbf{gap}} &

       $\textbf{\sigma }$ ($\textbf{\epsilon}$) &\textbf{AUC}  &
       \makecell{\textbf{Privacy} \\ \textbf{Leakage}}
       & \makecell{\textbf{Train} \\ \textbf{gap}} &
       \textbf{AUC}  & \makecell{\textbf{Train} \\ \textbf{gap}} 
       
       \\\hline

       \multirow{8}{*}{CoLA} &--   & 60.8 &.0811 & 15.8  & 61.4 & 15.8 & --   & 52.2 & .0077 & -3.86&52.7&-3.86\\
          & $.024 (10^6)$ & 52.0 & .0154& 2.31 & 51.6 &2.31 & $.014 (10^6)$  &48.7 &.0039 & 3.08 &50.3&3.08\\
          & $.066 (10^5)$  &  51.8 & .0077& 1.93  & 51.2 &1.93 & $.043 (10^5)$   & 48.8  &.0039 &2.70  &49.3&2.70\\
          & $ .112 (10^4)$  & 51.6  & .0058& 1.93 & 51.0 &  1.93& $ .092 (10^4)$   & 48.7  & .0039& 2.70 &49.5&2.70\\
          & $ .173 (10^3)$  &  51.5 &.0077 & 1.93 & 51.0 & 1.93 & $ .140 (10^3)$  & 48.7  &.0019 &  3.08&48.8&3.08\\
          & $.347 (10^2) $  & 51.4  & .0077& 1.93 & 50.6 & 1.93  & $ .278 (10^2)$   &  48.6 & -.0058& 3.86&49.4& 3.86\\
          & $ .747 (10^1)$  & 51.4  & .0058& 2.31  & 50.4 & 2.31&$ .615 (10^1)$   & 48.0  & -.0212& 5.40&47.4& 5.40\\
          & $ 3.06 (10^0)$  &  51.4 &.0039 & 2.31 & 50.5 & 2.31& $1.91 (10^0) $   & 46.6  & -.0193& 7.33&46.7& 7.33 \\
         
         \cline{2-13}
          \multirow{8}{*}{IMDb} & --  & 56.7 &.0496 & 6.5  &58.2 & 6.5 & -- & 54.6 & .0303& -0.04 &57.4&-0.04 \\
           & $.019 (10^6)$   & 55.2 &.0320 & 9.68 & 54.4&9.6   &$.016 (10^6)$  &  52.7 & .0232& 6.21&52.9&6.21  \\

          & $.052 (10^5)$  & 54.5  & .0259& 6.94 & 54.0 & 6.94 &$.072 (10^5)$  & 52.4  &.0127 & 5.52&52.3& 5.52 \\
           & $ .089 (10^4)$   & 54.2  & .0232& 6.40  & 53.9 & 6.40 &$ .086 (10^4)$   & 52.3  &.0132 &5.13&52.2& 5.13 \\
           & $ .121 (10^3)$  &  53.9 & .0240& 4.49 & 53.9 & 4.49  &$ .116 (10^3)$   &  52.1 & .0166& 3.76 &51.9&3.76\\
           & $ .226 (10^2)$   & 53.7  &.0289 & 3.12 & 53.4 & 3.12 &$.216 (10^2) $   & 51.2  & .0144& 3.86 &51.6&3.86\\
          & $ .472 (10^1)$  & 53.1  & .0240& 4.35 & 52.2 & 4.35  &$.460  (10^1)$ &51.0  & .0071&  3.52&51.2&3.52\\
           & $1.01 (10^0) $   &  52.6 &.0120 & 3.61 & 51.5 & 3.61 &$.993 (10^0) $   &  51.5 & .0115& 4.15&51.8&4.15 \\

          \cline{2-13}
          \multirow{8}{*}{SST2} & -- & 51.1 &.0105 & -0.06 & 51.3&-.06 & --   & 50.3 &.0027 & -2.0&51.4& -2.0 \\
           & $.034 (10^6)$   & 50.6 &.0037 & -1.16 & 50.4& -1.16 & $.042 (10^6)$  & 49.6 &-.0047 & -0.22 &49.5&-0.22 \\

           & $.072 (10^5)$ & 50.2  & .0011& -0.67  & 50.5& -0.67 & $.078 (10^5)$  & 49.5  & -.0047& -0.22 &49.5&-0.22\\
           & $ .096 (10^4)$  & 50.0  & .0000& -0.06  & 50.4 & -0.06 & $ .101 (10^4)$   &49.4   & -.0047&-0.22 &49.5&-0.22 \\
           &$ .127 (10^3)$   & 49.7  & -.0019& -0.93 & 49.9  &  -0.93 & $ .137 (10^3)$ & 49.4  & -.0047& -0.22 &49.5&-0.22\\
           & $.234  (10^2)$  & 49.7  & -.0050&-2.06  & 49.6 &  -2.06 & $ .254 (10^2)$  &  49.4 & -.0047& -0.22 &49.5&-0.22\\
           & $.458 (10^1) $   & 49.7  & -.0047&  -2.19 & 49.6   &  -2.19\ &$ .478 (10^1)$   &49.4   & -.0047& -0.22 &49.5&-0.22\\
           & $ .957 (10^0)$  &  49.8 & .0047& -2.22 & 49.6  & -2.22 & $ 1.00 (10^0)$   &   49.4&-.0047 &  -0.22&49.5&-0.22\\

          \hline
        \end{tabular}
       \caption{\textmd{Attack metrics for MIA and MIA-R.}} 
       \label{tab:mias}
   \end{table*}

       \begin{table}[ht!]
    \setlength{\tabcolsep}{4.5pt}
       \centering
       \small
       \begin{tabular}{ccc|cc}
       \hline
       &\multicolumn{2}{c|}{\textbf{GPT2}}&\multicolumn{2}{c}{\textbf{BERT}} \\

       \textbf{Dataset} &
       $\textbf{\sigma }$ ($\textbf{\epsilon}$) &
       
       \textbf{AUC}  &

       $\textbf{\sigma }$ ($\textbf{\epsilon}$) &
       \textbf{AUC} 
       
       \\\hline

       \multirow{8}{*}{CoLA} &--   &  17.2 &--   &32.6 \\
          & $.024 (10^6)$ &  17.3 & $.014 (10^6)$  & 52.8\\
          & $.066 (10^5)$  &16.8  &$.043 (10^5)$   & 57.2\\
          & $ .112 (10^4)$  & 16.4 & $ .092 (10^4)$   &58.7\\
          & $ .173 (10^3)$  &  16.2  & $ .140 (10^3)$  & 59.4 \\
          & $.347 (10^2) $  & 15.7 & $ .278 (10^2)$   & 60.7 \\
          & $ .747 (10^1)$  & 15.4 &$ .615 (10^1)$   &62.7  \\
          & $ 3.06 (10^0)$  &  15.6 & $1.91 (10^0) $  & 64.9   \\
         
         \cline{2-5}
          \multirow{8}{*}{IMDb} & --  &52.6  & -- & 50.6 \\
           & $.019 (10^6)$   &52.9   &$.016 (10^6)$  &  50.5 \\

          & $.052 (10^5)$  & 52.2 &$.047 (10^5)$  &50.5 \\
           & $ .089 (10^4)$ & 51.9  &$ .086 (10^4)$   & 50.4\\
           & $ .121 (10^3)$ & 51.7&$ .116 (10^3)$  &50.2 \\
           & $ .226 (10^2)$   &   51.5 &$.216 (10^2) $   &49.7 \\
          & $ .472 (10^1)$  &  51.2  &$.460 (10^1) $ & 49.5\\
           & $1.01  (10^0)$   &50.6  &$.993  (10^0)$   & 50.4\\

          \cline{2-5}
          \multirow{8}{*}{SST2} & -- & 53.7& --   &64.3 \\
           & $.034 (10^6)$   & 60.0 & $.042 (10^6)$  & 61.7 \\

           & $.072 (10^5)$ & 60.3 & $.078 (10^5)$  & 60.5\\
           & $ .096 (10^4)$  & 59.1  & $ .101 (10^4)$   &58.7 \\
           &$ .127 (10^3)$  & 58.5& $ .137 (10^3)$ & 56.9\\
           & $.234 (10^2)$  & 56.5& $ .254 (10^2)$  & 53.2\\
           & $.458  (10^1)$   & 53.4 &$ .478 (10^1)$   &52.6 \\
           & $ .957 (10^0)$  & 49.7 & $ 1.00 (10^0)$   & 52.0 \\

          \hline
        \end{tabular}
       \caption{\textmd{AUC for MIA-N.}} 
       \label{tab:mian}
   \end{table}

   \begin{table*}[ht!]
     \setlength{\tabcolsep}{4.5pt}
        \centering
        \small
        \begin{tabular}{cccccc|ccccc}
        \hline
        &\multicolumn{5}{c|}{\textbf{GPT2}}&\multicolumn{5}{c}{\textbf{BERT}} \\

        \textbf{Dataset} &
        $\textbf{\sigma }$ ($\textbf{\epsilon}$) &
        
        \textbf{Jaccard} $\uparrow$  &
        \makecell{\textbf{Cos.} $\uparrow$ \\ \textbf{sim.}}
        & \textbf{Meteor} $\uparrow$  & \textbf{RougeL} $\uparrow$ &

        $\textbf{\sigma }$ ($\textbf{\epsilon}$) &
        \textbf{Jaccard} $\uparrow$  &
        \makecell{\textbf{Cos.} \\ \textbf{sim.}} $\uparrow$
        & \textbf{Meteor} $\uparrow$ & \textbf{RougeL} $\uparrow$ 
        
        \\\hline

        \multirow{8}{*}{CoLA} &--   &.6289 & .2135 & .7430 &   .6882 &--   &.5992& .4895&.6281  &.6972  \\
           & $.024 (10^6)$ &  .0149 & .0997&   .0192  & .0286& $.014 (10^6)$  & .0155&.1310  & .0352  &.0257\\
           & $.066 (10^5)$  & .0004&.0965 &.0048 &.0022 &$.043 (10^5)$   & .0047   & .1134 & .0270  &.0090\\
           & $ .112 (10^4)$  & .0001  & .0967&   .0045 & .0017 & $ .092 (10^4)$   & .0004   &.1102  &.0237   &.0011\\
           & $ .173 (10^3)$  &   .0001 &.0966 & .0045&.0017  & $ .140 (10^3)$  &  .0002 &.1103 & .0236& .0009\\
           & $.347  (10^2)$  & .0001  & .0966&  .0045    & .0017 & $ .278 (10^2)$   &  .0002  & .1103 &.0236   &.0009\\
           & $ .747 (10^1)$  & .0001  & .0966&  .0045  & .0017  &$ .615 (10^1)$   &  .0002  & .1103 &.0236    &.0009\\
           & $ 3.06 (10^0)$  & .0001  & .0966&  .0045   & .0017  & $1.91  (10^0)$   &  .0002  & .1103 &.0236   &.0009 \\
          
          \cline{2-11}
           \multirow{8}{*}{IMDb} & --  & .7561 & .4352&.8538  & .8175& -- & .7095&.7360 & .7709 &.8374\\
            & $.019 (10^6)$   & .0102  &.1898 & .0260   & .0199 &$.016 (10^6)$  &  .0174  &  .1292&  .0521 & .0198 \\

           & $.052 (10^5)$  & .0016&  .1775 &.0174 &.0055  &$.047 (10^5)$  &.0028&.1020  &  .0410 & .0042\\
            & $ .089 (10^4)$ &  .0011 & .1764&.0168  &.0048&$ .086 (10^4)$   &.0021 &.1005  &   .0403&.0032 \\
            & $ .121 (10^3)$ &.0011 & .1764&.0168  &.0047  &$ .116 (10^3)$  &.0020& .1002& .0403&.0032\\
            & $ .226 (10^2)$   &  .0011 & .1764&.0168  &.0046  &$.216  (10^2)$   &.0020 &.1002 & .0403& .0032\\
           & $ .472 (10^1)$  &   .0010 & .1760&.0167 &.0046  &$.460  (10^1)$ & .0020   &.1002  &  .0403 &.0031\\
            & $1.01 (10^0) $   & .0010 & .1760&.0167 &.0046  &$.993  (10^0)$   &  .0020  & .1002 &  .0403& .0031\\

           \cline{2-11}
           \multirow{8}{*}{SST2} & -- &.6549 &.3118 & .8171 & .7277& --   &.6531 &.5167 & .6805 &.7498\\
            & $.034 (10^6)$   & .0031&.1017 & .0098 & .0063& $.042 (10^6)$  &.0035& .1284&.0341 & .0063 \\

            & $.072 (10^5)$ & .0003  &.1015 &.0063    & .0014 & $.078 (10^5)$  & .0010   &  .1266&  .0326 &.0023\\
            & $ .096 (10^4)$  &  .0001 & .1012& .0062  & .0009 & $ .101 (10^4)$   &.0008& .1273 &  .0323 &.0018 \\
            &$ .127 (10^3)$  & .0001 & .1013&.0062 & .0009 & $ .137 (10^3)$ &   .0008 &.1270 &.0323 & .0018\\
            & $.234 (10^2) $  & .0001 & .1012& .0062 & .0009& $ .254 (10^2)$  &.0008    & .1267 &  .0324  &.0018\\
            & $.458 (10^1) $   & .0001  &.1012 &.0062   & .0009 &$ .478 (10^1)$   &.0008&  .1267& .0324 &.0018\\
            & $ .957 (10^0)$  & .0001  & .1012&.0062   & .0009 & $ 1.00 (10^0)$   &.0008    &.1267  &  .0324  &.0018 \\

           \hline
         \end{tabular}
        \caption{\textmd{Decepticons attack: reconstruction metrics against non private and private models. The attack collapses after some levels of noise.} } 
        \label{tab:decepticons_metrics}
    \end{table*}

    \paragraph{MIA}
    We start the privacy analysis by comparing how susceptible a model trained without DP is to models trained with the Gaussian mechanism against the membership inference attacks (MIA) and with reference models (MIA-R) from \cite{Shokri2016MembershipIA}  and \cite{ye-etal:2022:CCS}. These are generalist attacks, since they only require access to the logits of the target model.

    We first deployed the attacks against non private models trained for the same number of epochs as in Utility above. 
    We find that the general performance of the attacks is poor across the board, in line with \cite{duan2024membershipinferenceattackswork}, so we omit these results.
    An explanation is the small gap between test and training set. 
    However, when we increase the number of epochs or decrease the batch size to deliberately try to overfit the models (measured by the train-test gap), we start seeing bigger AUC scores, as shown in Table~\ref{tab:mias}.

    For both MIA flavours, the only setting for which the AUC reached 60\% was for GPT2 trained with CoLA sentences. For a comparison, \cite{mireshghallah-etal-2022-quantifying}  conducted a study on MIA against masked language models (MLM), and achieved an AUC ranging from 66\% to 90\% at datapoint level. 
    Not surprisingly, this was also the case which the model was overfitted the most. Opposite to CoLA, the AUC for SST2 sentences stays around random chance for all settings. Last, IMDb sits in the middle.

    The privacy leakage metric broadly shows the same patterns, also revealing a sharp drop in the attack success when DP is applied. With very few exceptions, its success remains below 1\%.  Most importantly, the relationship between $\sigma$ and privacy leakage (and likewise AUC) is quite erratic, compromising attempts to find any pattern.  We also observe that a very small noise (the smallest across our experiments) is enough to almost neutralise the attack in almost all settings, severely limiting it usefulness for calibrating privacy. 
    Other studies have also pointed that MIAs struggle to achieve high success against non-overfitted models \cite{8962136}.

    In realistic scenarios, standard regularisation techniques as dropout or early stopping are efficient to reduce the training-test gap. Moreover, fine tuning is performed for a few epochs only.

    We also report in Table~\ref{tab:mian} results for the MIA-N attack, which does not require shadow models. Instead, neighbouring samples are artificially generated. Contrary to the other flavours, this attack is much more successful against BERT trained on SST2, and slightly more successful against GPT2. For IMDb, however, it struggles, performing essentially at chance across the range of $\sigma$. 
    The results for CoLA are particularly odd.  Under GPT2 the AUC is $< 20$ (much worse than chance), indicating the attack is consistently making the wrong prediction about membership.  Under BERT, the relationship of AUC with noise is the reverse of what is expected. This may be connected to the imbalanced nature of CoLA, but it is unclear.  In any case, MIA-N is not suitable for calibration.

    In summary, there is no MIA that works well across all  these popular datasets, and it consequently does not seem suitable for calibration.

    \paragraph{Gradient-based reconstructions} We evaluate two attacks which reconstruct data from shared gradients in a federated learning training. The first, Decepticons, was proposed to work against Transformers, while the second, Lamp, attacks BERT. Hence, they are more specific than MIA.

    \textit{Decepticons attack} We first report several metrics for the Decepticons attack against non-private models in Table~\ref{tab:decepticons_metrics}. We focus on ROUGE-L in particular, as this is commonly used in evaluations of reconstruction attacks \cite{deng-etal-2021-tag-gradient, NEURIPS2022_32375260}, but include the others for alternative perspectives.
    In general, the attack is very successful at reconstructing the input from the gradients. Surprisingly, longer texts (IMDb) are better reconstructed than short ones (SST2 and CoLA). This observations holds across all metrics.

    GPT2 seems slightly more susceptible to noise compared to BERT based on the trends in the metrics. For example, in the CoLA dataset, Jaccard and Meteor for GPT2 drop faster than BERT as noise increases.

    However, the moment that noise is injected to the gradients, reconstruction is strongly affected. As noise increases, performance drops in a very consistent way, until (in all experimental combinations) the results bottom out. In the metrics, this occurs when noise multiplier $\sigma$ surpasses 0.1. Both \cite{ponomareva-etal:2023:JAIR} and \cite{hayes-etal:2024:NeurIPS} remark that larger $\epsilon$ is sufficient for protecting against reconstruction, so it is perhaps not unexpected that a minimum level of reconstruction might be reached for stronger noise.

\begin{table}[h!]
     \setlength{\tabcolsep}{2pt}
        \centering
        \small
        \begin{tabular}{cccccc}
        \hline
        
        \textbf{Dataset} &
        $\textbf{\sigma }$ ($\textbf{\epsilon}$) &
        
        \textbf{Jaccard} $\uparrow$ &
        \makecell{\textbf{Cos.} \\ \textbf{sim.}}$\uparrow$
        & \textbf{Meteor}$\uparrow$  & \textbf{RougeL} $\uparrow$
        
        \\\hline

        \multirow{8}{*}{CoLA} &--  &.5508 &.0801 &.6303 &.6115   \\
           & $.014 (10^6)$  & .4363 &.0740 &.4903 & .5105 \\
           &$.043 (10^5)$ & .2916 &.0686&.3618 & .3988\\
           & $ .092 (10^4)$ &.0817  &.0507 &.1164   & .1435\\
           & $ .140 (10^3)$ &  .0170&.0459 &.0307 &.0482 \\
           & $ .278 (10^2)$  & .0013 & .0369  & .0037 & .0030\\
           &$ .615 (10^1)$  & .0002 &  .0390 & .0017   & .0013\\
           & $1.91  (10^0)$ & .0003 & .0351  & .0020   &.0016 \\
          
          \cline{2-6}
           \multirow{8}{*}{IMDb} & -- &.0571& .2498&.1036  &.0693 \\
            &$.016 (10^6)$ &  .0432& .2321&  .0856 &.0546 \\

           &$.047  (10^5)$    & .0404 &  .2255 &.0775 &.0499 \\
            &$ .086 (10^4)$   & .0386 &.2208   &  .0758  &.0475 \\
             &$ .116 (10^3)$  & .0367 &  .2212 & .0701   &.0447 \\
            &$.216 (10^2) $    &  .0364 &.2150 & .0674 &.0466\\
            &$.460  (10^1)$  & .0361 &  .2121 &  .0663  &.0438 \\
            &$.993 (10^0) $ & .0330 & .2109  & .0625   &.0422\\

           \cline{2-6}
           \multirow{8}{*}{SST2} & --   &.6835 & .0853&.6011 &.6524\\
            &$.042 (10^6)$  & .4408& .0704& .3831&.4577 \\

            & $.078 (10^5)$ &.3111  & .0644  & .2457   & .3329\\
             & $ .101 (10^4)$ & .2393 &  .0605 & .1901  & .2677\\
             & $ .137 (10^3)$ &.1617 & .0538& .1362 &.1829 \\
            & $ .254 (10^2)$ & .0269 &  .0437 & .0211    &.0271 \\
            &$ .478 (10^1)$& .0003 &  .0446 &  .0068 & .0032\\
            & $ 1.00 (10^0)$ &.0003  & .0384  &  .0092  &.0028 \\

           \hline
         \end{tabular}
        \caption{\textmd{LAMP attack: reconstruction metrics against non-private and private BERT.}}
        \label{tab:LAMP_metrics}
    \end{table}

    \textit{LAMP attack} We report in Table~\ref{tab:LAMP_metrics} reconstruction metrics for the LAMP attack against BERT. The evaluation is conducted for both non-private and private models with different noise levels $\sigma$.

    It is in general a more successful attack than Decepticons.
    As with Decepticons, it also shows a smooth relationship between level of noise and metric value.
    In contrast to Decepticons, this attack shows a less dramatic drop-off from non-private to private, and does not bottom out, with even the protection against reconstruction improving for even the smallest privacy budgets.

    Regarding datasets, CoLA and SST2 show a sharper decline in reconstruction quality compared to IMDb, which exhibits a more gradual or stable decrease. However, such gradual decrease is shadowed by the fact that LAMP struggles to reconstruct IMDb texts from the non private model. A likely reason is its substantially longer texts.

     In terms of metric behaviour, then, this attack seems to be particularly suitable for calibrating noise. However, it only applies to BERT models, which is something of a limitation.

\section{Directional privacy with VMF}
\label{sec:privacy_model}

Following our conclusions about empirical privacy calibration in Sec~\ref{sec:calibration}, in this section we apply them to comparing the commonly used Gaussian mechanism of DP-SGD to one that falls within a generalisation of DP known as metric DP, using the Von-Mises Fisher (VMF) mechanism.

\subsection{Metric DP}

\paragraph{Background}
    \cite{chatzikokolakis-etal:2013:PETS} introduced a generalisation of DP, now usually referred to as metric DP, where the indistinguishability requirement depends on an arbitrary notion of distance; standard DP is then the specific instance of the Hamming distance. 


More formally, for $\varepsilon{>}0$, a mechanism ${\mathcal M}$ on an (input) metric space $(S, d)$ satisfies $\varepsilon d$-privacy, if for all $s, s'\in S$ and $Z \subseteq \mathcal{Z}$, where $S$ is a set of inputs, and $\mathcal{Z}$ is the set of outputs, 

\vspace{-0.3cm}

    \begin{equation*}\label{ddp_eqn}
        Pr ({\mathcal M}(s))[Z] \leq e^{\varepsilon d(s, s')}\times Pr({\mathcal M}(s'))[Z]~ ,
    \end{equation*}
   
\noindent
where $Pr ({\mathcal M}(s))[Z]$ means the probability that the output of applying $\mathcal{M}$ to the input $s$ is in $Z$.

When two inputs $s,s'$ differ by the amount $d(s,s')$ (which is different from the traditional DP, where inputs are said to be neighbouring if they differ by one element), ${\mathcal M}$ makes them indistinguishable proportional to $e^{\varepsilon d(s, s')}$. Intuitively, the closer they are, the harder it is to distinguish them.

    The first application was in geo-location privacy \cite{andres2013geo}, in which a user provided an approximate location to receive services, and indistinguishability was related to Euclidean distance. In NLP, \cite{fernandes-etal:2019:POST} proposed a \dpriv mechanism for authorship privacy based on Euclidean distance on word embeddings, with other work following \cite{10.1145/3336191.3371856,yue-etal-2021-differential,SilvaCarvalho2023}.

\begin{figure}[!th]{\textwidth 0mm}
		\centering
		\includegraphics[width=0.35\textwidth]{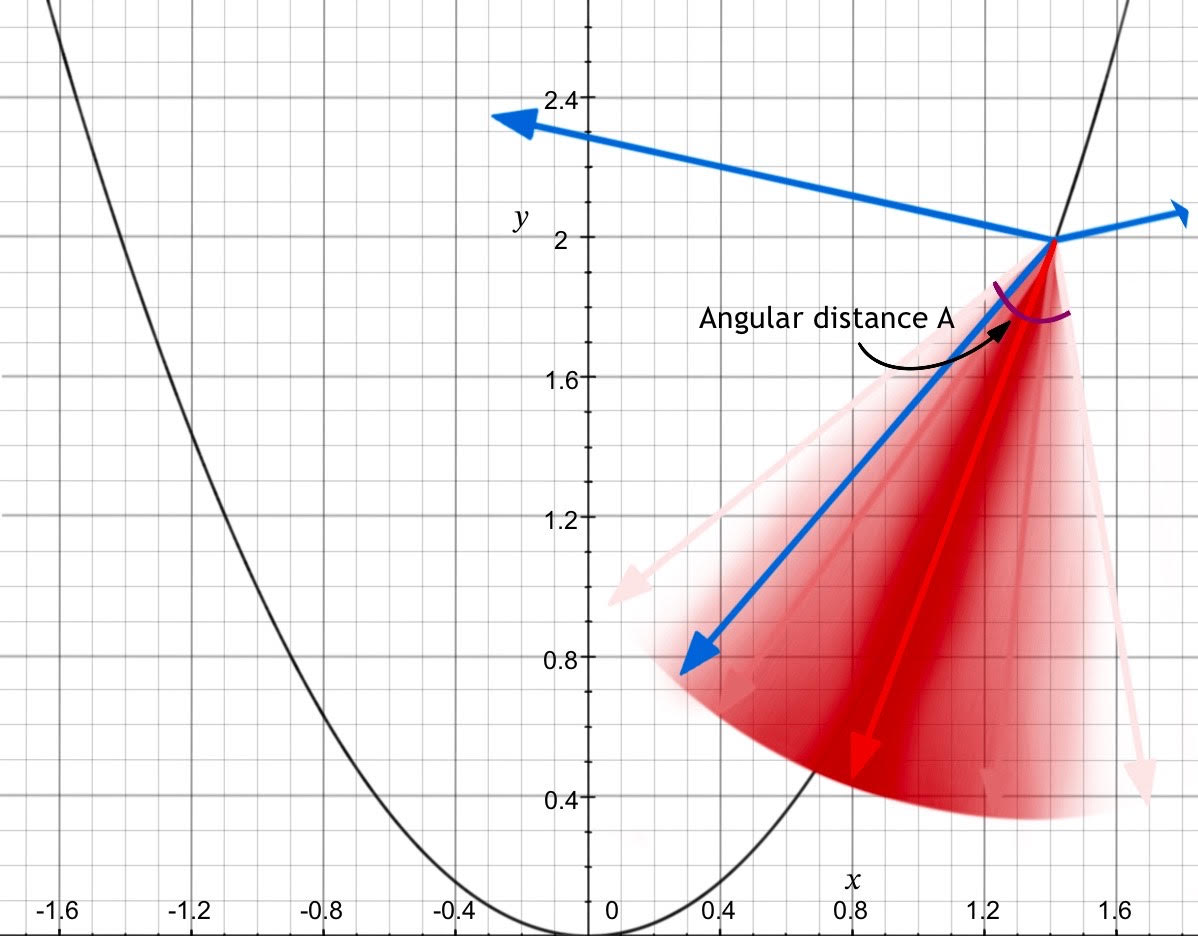}
		
		\caption{Illustration highlighting how gradients are perturbed with Gaussian noise versus VMF noise. The dark red arrow is the unperturbed gradient, and lighter red arrows are perturbations of angular distance $A$ that are more likely when close to the unperturbed gradient, as under the VMF mechanism. The blue arrows represent isotropic noise added to the gradient by DP-SGD.}\label{f1406-a}
	\end{figure}

    Gradient descent optimises the search for parameter selection that minimises the loss. Thus an alternative method of perturbing the gradients is to use randomisation that is based on perturbing the angle of deviation from the original gradient.
    To give some intuition, Figure~\ref{f1406-a} illustrates how a gradient of a convex curve can be perturbed, leading to a perturbation of the descents. 

    Given two vectors $v, v'$ in $\mathbb{R}^K$, the angular distance between them is  $d_\theta(v, v')= \frac{\arccos{v^Tv'}}{\| v\| \| v'\|}$. When $v,v'$ are, for example, vectors on the unit $K$-dimensional sphere, then $d_\theta$ becomes a metric. Following  \cite{weggenmann-kerschbaum:2021:CCS}, we can use this to define \emph{directional privacy}.

    \begin{definition}\label{d1614}
    (Directional Privacy) \cite{weggenmann-kerschbaum:2021:CCS} Let $\epsilon{>}0$. A mechanism ${\mathcal M}$ on $\mathbb{R}^K$ satisfies $\varepsilon d_\theta$-privacy, if for all $v, v'$ and $Z \subseteq \textrm{supp}\mathcal{M}$, 
    \[
    Pr({\mathcal M}(v))[Z] \leq e^{\varepsilon d_\theta(v, v')}\times Pr({\mathcal M}(v'))[Z]~.
    \]
    \end{definition}
    
    
    Definition~\ref{d1614} says that when the mechanism ${\mathcal M}$ perturbs the vectors $v, v'$, the probabilities that the perturbed vectors lie within a (measurable) set $Z$ differ by a factor of $e^{\varepsilon d_\theta(v, v')}$. This means that the smaller the angular distance between the vectors $v,v'$ the more indistinguishable they will be.

\paragraph{Metric DP for SGD}
In the following, we provide some proofs that the use of the VMF mechanism does provide metric DP guarantees, extending from proofs provided where it was first introduced to our in training machine learning architectures.

    \cite{weggenmann-kerschbaum:2021:CCS} introduced the von Mises-Fisher (VMF) mechanism derived from the VMF distribution that perturbs an input vector $x$ on the $K$-dimensional unit sphere.
    \begin{definition}\label{d1526}
    The VMF mechanism on the $K$-dimensional unit sphere is given by the density function:
    \[
    \VM(\varepsilon, x)(y) ~ = ~ C_K(\varepsilon) e^{\varepsilon x^T y}~,
    \]
    where $\varepsilon >0$ and $C_K(\varepsilon)$ is the normalisation factor.
    \end{definition}
    
    \cite{weggenmann-kerschbaum:2021:CCS} proved that the VMF mechanism satisfies $\varepsilon d_\theta$-privacy. 
    
    They also provide an analysis of the expected error of the VMF\footnote{See \cite{weggenmann-kerschbaum:2021:CCS} for full details.}  as well as sampling methods which we use later in our experiments. Importantly, the following also holds. 
    
    \begin{theorem}[\cite{weggenmann-kerschbaum:2021:CCS}]\label{thm:d2_priv}
    Let $\epsilon > 0$ and denote by $\mathbb{S}^{K-1}$ the unit sphere in $K$ dimensions. Then the VMF mechanism on $\mathbb{S}^{K-1}$ satisfies $\epsilon d_2$-privacy where $d_2$ is the Euclidean metric. That is,
    \[
    	\VM(\epsilon, x)(Y) ~\leq e^{\epsilon d_2(x, x')} \VM(\epsilon, x')(Y)~,
    \]
    for all $x, x' \in \mathbb{S}^{K-1}$ and all (measurable) $Y \subseteq \mathbb{S}^{K-1}$.
    \end{theorem}

     In our application, we will need to apply the basic mechanism $\VM$ to more complex data representations, namely where a point is a represented as convex sum of $m$ orthogonal vectors in $n$ dimensions. The standard method for doing this is to apply $m$ independent applications of the mechanism (in this case $\VM$); the differential privacy guarantee is then parametrised by $m$ as follows.
    
    \begin{corollary}[\cite{dwork-roth:2014}]\label{c1326}
        Let $\VM$ be the mechanism defined in Definition~\ref{d1526}. Let $v, v'$ be two vectors on the unit sphere, where $v= \lambda_1u_1 + \dots \lambda_k u_m$ and  $v'= \lambda'_1u'_1 + \dots \lambda'_k u'_m$, where $u_i, u'_i$ for $1\leq i \leq m$ are vectors on the unit sphere, and $|\lambda_i|, |\lambda'_i| \leq 1$. Define $\VM^*$ to be the mechanism that applies $\VM$ independently to each of the $u_i/u'_i$ to produce  random vectors distributed respectively as: $\VM^*(\epsilon, v), \VM^*(\epsilon, v')$. Then 
    
        \[
         \VM^*(\epsilon, v)(Y) ~\leq e^{2\epsilon \sqrt{m}} \VM(\epsilon, v')(Y)~.
        \]
    \end{corollary}

During training, standard DP-SGD calculates per-sample gradients $g_t(x_i)$ w.r.t the loss function. They are then clipped to an arbitrary length and perturbed using the Gaussian distribution. \cite{faustini2023directional} proposed clipping the gradients to 1 and replaced the Gaussian with VMF, referring to this SGD variant as \DirDP. 
For the VMF mechanism, \cite{weggenmann-kerschbaum:2021:CCS} observed that while Laplace and Gaussian are typical mechanisms used in Euclidean space, ``the VMF distribution can be seen as [a] natural counterpart on the sphere,'' specifically the unit hypersphere.  
The VMF distribution is defined by mean direction $\mu$ and concentration parameter $\kappa$; like the use of Laplace in DP, $\mu$ is set to zero, and $\kappa$ reflects the magnitude of the perturbation.

\Alg{alg:sgd2} depicts \DirDP. As with the original DP-SGD, lines 7 and 9 clip the gradients $g_t(x_i)$ to norm 1 and then noise is added. Instead of \emph{adding} a noisy vector to the gradients, the algorithm generates a noisy gradient directly. It follows \cite{weggenmann-etal:2022:WWW}, in which the larger the $\kappa$, the more the direction of the gradients is preserved, leading to better utility.


\begin{algorithm}[!th]
\small
\caption{\DirDP with VMF noise}\label{alg:sgd2}
\begin{algorithmic}[1]
\State \textbf{Input:} Samples $\{x_1,\ldots,x_N\}$, loss function $\mathcal{L}(\theta) = \frac{1}{N} \sum_i \mathcal{L}(\theta, x_i)$. Parameters: learning rate $\eta_t$, noise scale $\sigma$, group size $L$, {\color{blue}gradient norm bound $C = 1$.}
\State \textbf{Initialise} $\theta_0$ randomly
\For{$t \in T$}
   \Comment{Take a random batch}
   \State $L_t \gets $ random sample of $L$ indices from $1{\ldots}N$
   \For{$i \in L_t$}
         \Comment{Compute gradient vector}
   	\State $\mathbf{g}_t(x_i) \gets  \nabla_{\theta_t} \mathcal{L}(\theta_t, x_i)$
	\Comment{Scale gradient vector}
	{\color{blue}\State $\gbar_t(x_i) \gets \nicefrac{\mathbf{g}_t(x_i)}{\frac{\| \mathbf{g}_t(x_i)\|_2}{C} }$}
  \EndFor  
   \Comment{Add noise}
   \State {\color{blue}  $ \gtilde_t \gets \frac{1}{L} \sum_{i} \VM(\sigma, \gbar_t(x_i)) $}
   \Comment{Descent}
   \State $\theta_{t+1} \gets \theta_t - \eta_t \gtilde_t$
\EndFor
\State \textbf{Output} $\theta_T$ 
\end{algorithmic}
\end{algorithm}

To prove a DP guarantee, \Alg{alg:sgd2} is modified from the original DP-SGD (blue text) in three ways. First, we fix $C$ to $1$. Then (line 10), instead of clipping the gradients, we scale the gradients to the clip length (i.e. 1). Finally (line 13), instead of adding a noisy vector to the gradient, we generate a noisy gradient directly using the VMF mechanism.\footnote{This is because the VMF mechanism generates a new noisy vector based on its input.}

\Alg{alg:sgd2} satisfies $\epsilon$-DP and $\epsilon d_\theta$-privacy in terms of indistinguishability of batches used in the learning, viz that if two batches (composed of data points $x_i$) differ in only one point, then they are probabilistically indistinguishable.

\begin{theorem}\label{l1647}
Denote by $B = [v_1, \dots v_n]$ , and $B' = [v'_1, \dots v'_n]$ two batches of vectors (gradients). If batch $B'$ differs from $B$ in at most one component vector, then \Alg{alg:sgd2} satisfies $\sigma d_2$-privacy wrt.\ batches, namely that:

\begin{equation}\label{e1720}
Pr({\mathcal VM}(B) \in Z) \leq Pr({\mathcal VM}(B') \in Z) \times e^{\sigma d_2(B, B')} ~,
\end{equation}

 $Z$ is a (measurable) set of vectors, $Pr({\mathcal VM}(B) \in Z)$ is the probability that the output vector lies in $Z$ and (abusing notation) $d_2(B, B') = \max_{B \sim B'} d_2(v, v')$ is the maximum Euclidean distance between all pairs $v\in B, v'\in B'$.

\end{theorem}

Proof: see Appendix~\ref{proof:2}

Observe that \Alg{alg:sgd2} assumes we apply the $\VM$ mechanism to the whole gradient; as mentioned above, in our experiments we sometimes partition the $n$-dimensional space. We proceed though to prove a privacy guarantee assuming  \Alg{alg:sgd2} applies $\VM$ without partitioning. 

\begin{corollary}\label{cor:epsdp}
\Alg{alg:sgd2} satisfies $\epsilon$-DP wrt\ adjacent training sets $D, D'$.

\end{corollary}
Proof: see Appendix~\ref{proof:3}.

\begin{corollary}\label{cor:edpriv}
\Alg{alg:sgd2} satisfies $\epsilon d_\theta$-privacy.

\end{corollary}

Proof: see Appendix~\ref{proof:4}

Note that the epsilons in Cor~\ref{cor:epsdp} and Cor~\ref{cor:edpriv} are not comparable as they represent different notions of privacy.

We remark that by \emph{scaling} rather than clipping the gradients, we also protect against privacy leaks caused when the gradient length is less than $C$.  (Information may or may not be leaked by knowing the length of the gradient, but we remove this possibility by scaling rather than clipping.)

We note at this point that there are no existing theoretical relationships between the $\epsilon$ guarantees of standard DP and the distance-based ones of metric DP.  (\cite{mattern-etal-2022-limits}, in a work analysing word obfuscation with local DP, pointed out that metric DP changes the way $\epsilon$ is interpreted, which can lead to \textit{seemingly} smaller $\epsilon$, but metric DP does not change the privacy mechanism itself.) 
Aiming to compare DP-SGD and \DirDP is thus well suited to the empirical calibration framework of this paper.

\paragraph{Experimental setup}
We examine how models trained with VMF noise performs in comparison to Gaussian in preventing attacks at similar levels of utility. 
Based on our conclusions from Sec~\ref{sec:calibration}, for privacy calibration we use Decepticons and LAMP attacks rather than MIA. We present our principal results as plots of utility (accuracy or MCC) versus ROUGE-L for the two attacks.

\subsection{Results}\label{sec:util_results}

    We first observe briefly that utility results for VMF show broadly the same pattern as Gaussian (see plots in Appendix, Fig~\ref{fig:utility_plots}): utility tend to increase as noise diminishes (which for VMF is as $\kappa$ increases), with the same occasional erratic behaviour as noted for Gaussian.



    \begin{figure*}[h!]
    \centering
    \begin{tabular}{ccc}
               
        \subfloat[Tradeoffs for GPT2 on COLA]{\includegraphics[width=0.3\textwidth]{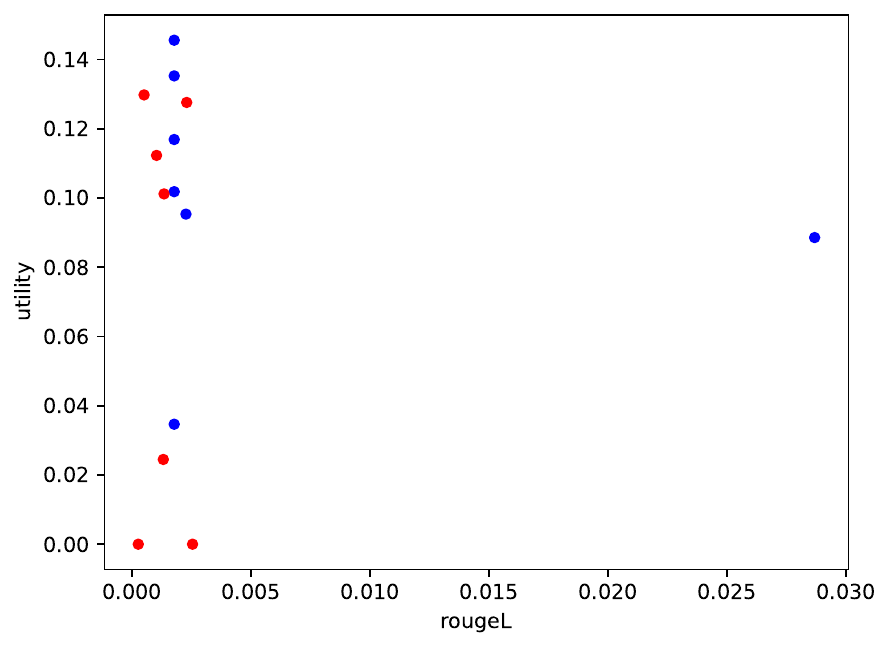}} &
         
        \subfloat[Tradeoffs for GPT2 on IMDb]{\includegraphics[width=0.3\textwidth]{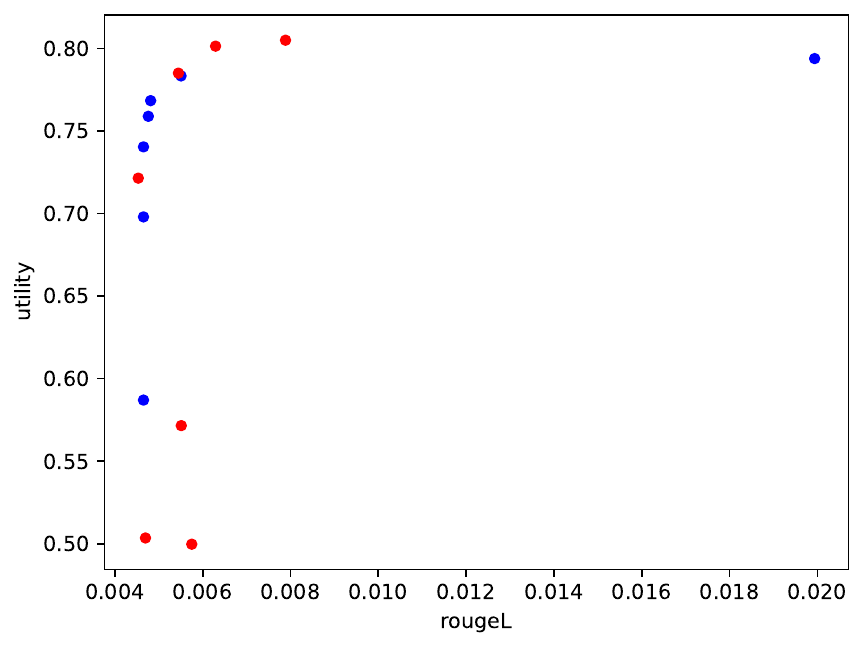}} &
        
        \subfloat[Tradeoffs for GPT2 on SST2]{\includegraphics[width=0.3\textwidth]{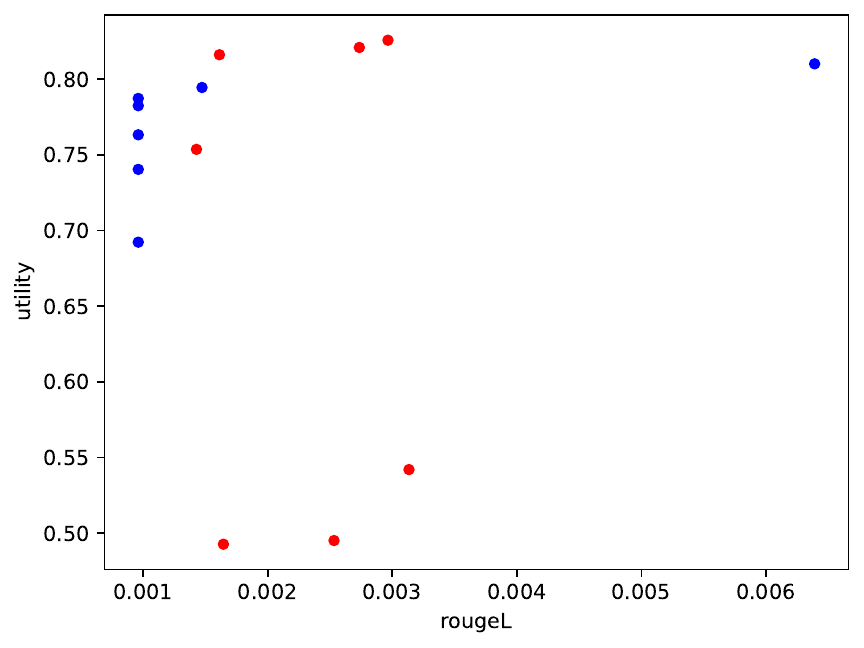}}  \\

        \subfloat[Tradeoffs for BERT on CoLA]{\includegraphics[width=0.3\textwidth]{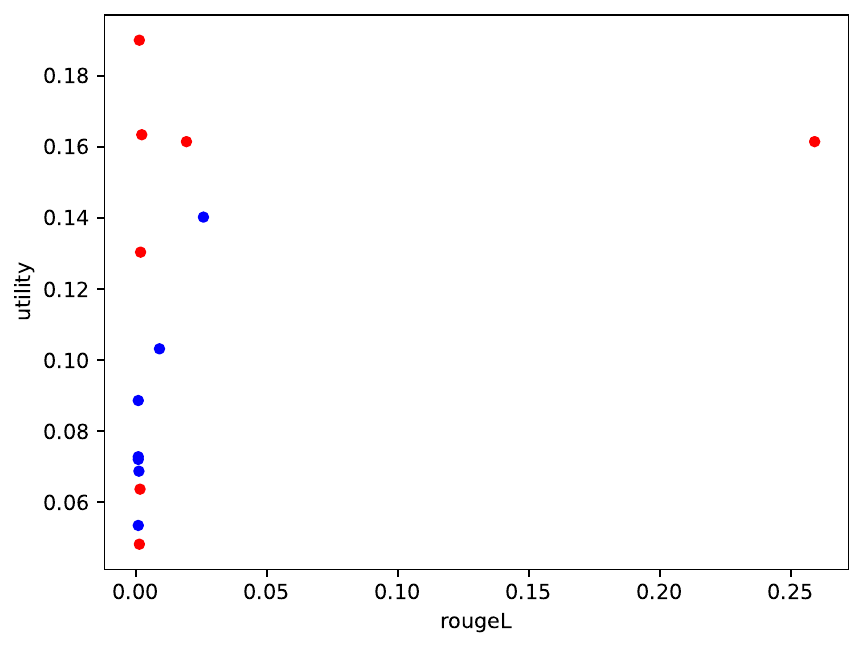}} &
        
        \subfloat[Tradeoffs for BERT on IMDb]{\includegraphics[width=0.3\textwidth]{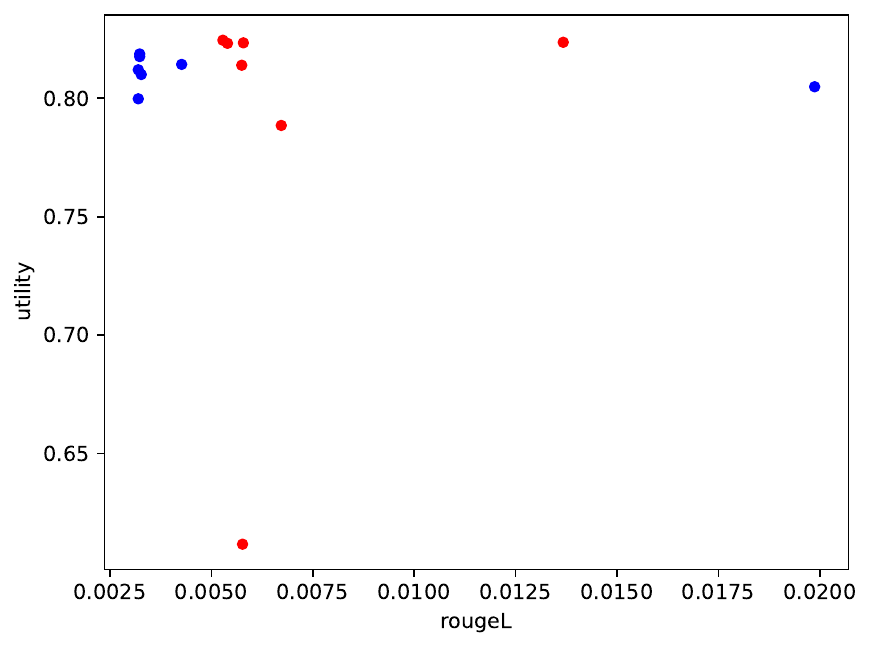}} & 

        \subfloat[Tradeoffs for BERT on SST2]{\includegraphics[width=0.3\textwidth]{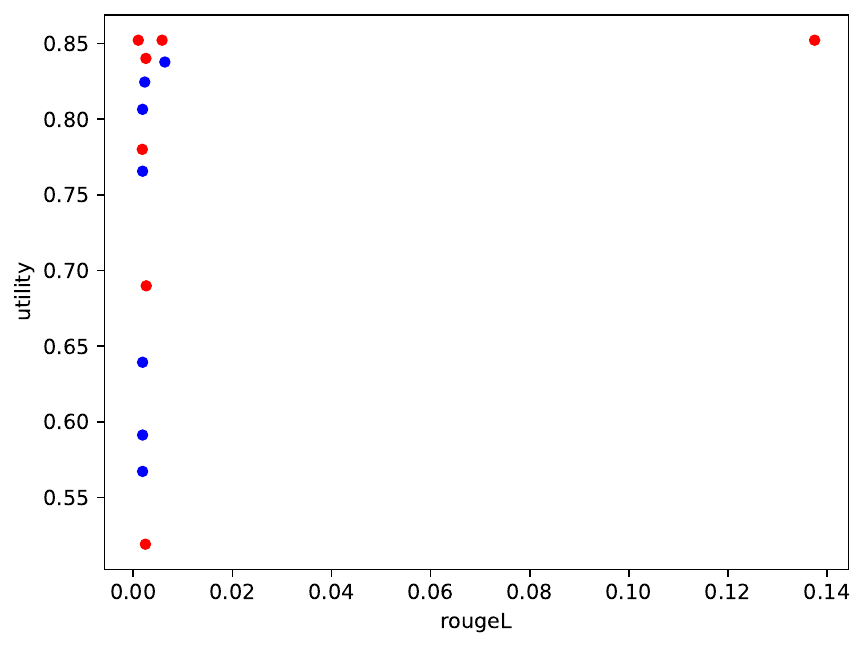}}
    \end{tabular}
    \caption{\textmd{Tradeoffs between privacy (ROUGE-L) and utility (accuracy or MCC), across the models and datasets under different privacy settings for the Decepticons attack.  Top left is best. }}\label{fig:decepticons_plots}
    \end{figure*}

    We plot in Figure \ref{fig:decepticons_plots} utility against ROUGE-L score of \textbf{Decepticons} for both mechanisms. Ideal dots are in the top left corner of each images. In such cases, the attack is prevented (small values for ROUGE-L) and the model trained with the same amount of noise ($\sigma$ for Gaussian or $\kappa$ for VMF) achieves a high utility.
    We can see that VMF outperforms Gaussian for BERT on CoLA and SST2 datasets --- the datasets with shorter texts --- and Gaussian outperforms VMF for BERT on IMDb. For the other scenarios, there is no clear winner.

    \begin{figure*}[h!]
    \centering
    \begin{tabular}{ccc}
               
        \subfloat[CoLA]{\includegraphics[width=0.3\textwidth]{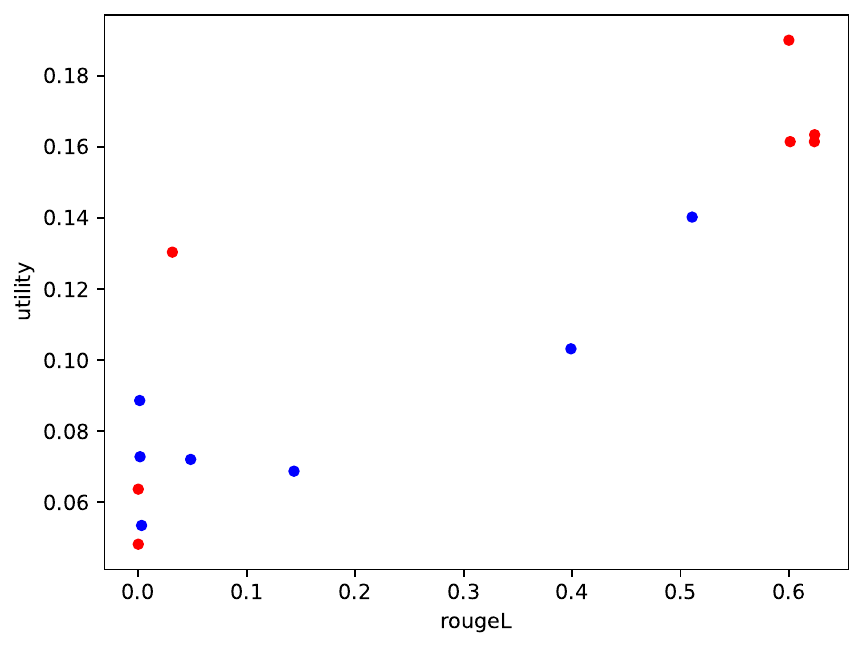}} &
        
        \subfloat[IMDb]{\includegraphics[width=0.3\textwidth]{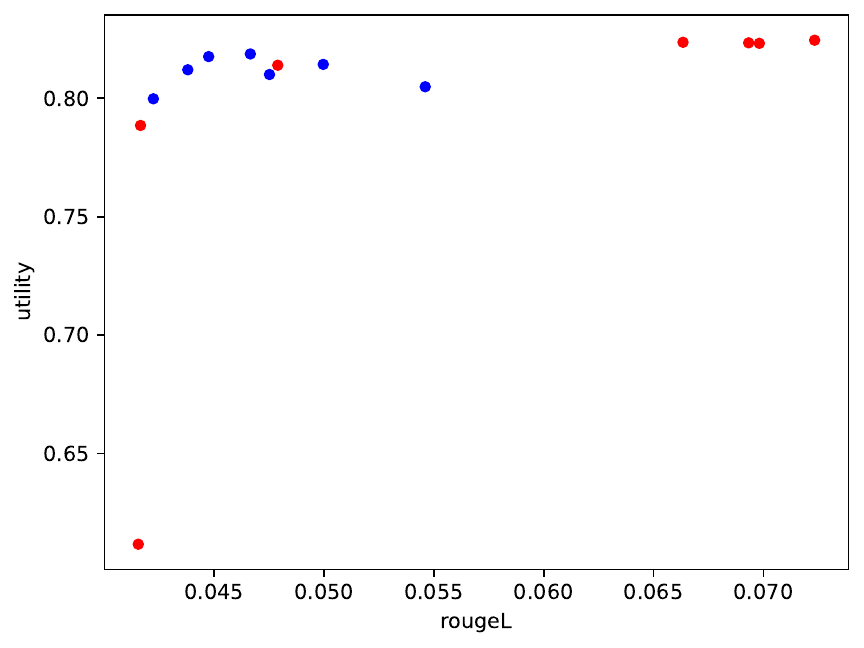}} &
    
        \subfloat[SST]{\includegraphics[width=0.3\textwidth]{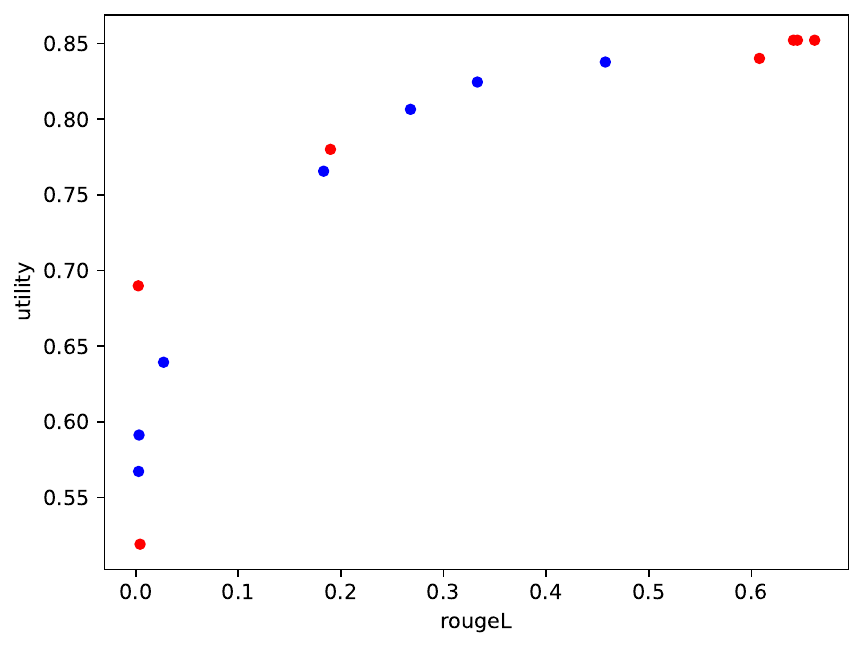}} 
    \end{tabular}
    \caption{\textmd{LAMP attack: reconstruction (ROUGE-L) against utility (accuracy or MCC) for Gaussian (blue dots) and VMF (red dots) noises. Top left is best.}}\label{fig:lamp}
    \end{figure*}

    We also assess how well the mechanisms protect against reconstruction by the \textbf{LAMP attack}. The plots in Figure~\ref{fig:lamp} show that it is possible to retain high utility for both mechanisms and preventing the attack in IMDb, although Gaussian is better in preventing reconstruction. For CoLA and SST2, the tradeoff is clearer --- the higher the utility, the better the reconstruction --- but no clear winning mechanism, although Gaussian does offer more intermediate ROUGE-L values.

    We have used ROUGE-L here as our principal calibration metric as it is widely used in reconstruction evaluation, and it seems broadly suitable at a coarse-grained level.  However, to our knowledge there is not yet any work on the extent to which the differences in ROUGE-L scores correlate with human judgements about reconstruction; contrast this with image processing, where MSE is used as an objective measure of distortion and SSIM as one that correlates with human perceptions \cite{scheliga-etal:2022:WACV}.
    We looked at a few examples.  
    Table~\ref{tab:lamp_examples_rouge} gives a few examples of reconstructed sentences that have the same ROUGE-L scores. Even when ROUGE-L scores are similar, humans might judge reconstructions to be of differing quality, in part because ROUGE-L does not distinguish among parts-of-speech (so reconstructing a determiner is as good as reconstructing a key noun). Additionally, reconstructions often contain non-words that give strong clues about the original but that ROUGE-L with its token matching does not pick up (e.g. ``who has seen my snorkel?'' reconstructed with noise as ``whokel scene has mynorkel?'').

\section{Conclusions}\label{sec:conclusions}

    We assessed the strengths and weaknesses of different attacks against NLP models in empirical privacy calibration. 
    Our results show that MIAs, a generic attack that can be deployed against a myriad of architectures, offers little help in calibrating for DP. However, gradient reconstruction attacks, which lie within the more specific framework of federated learning, can easily reconstruct short to medium texts if the gradients are not obfuscated. For the Decepticons attack, there seems to be a threshold for which reconstruction fails, whereas for LAMP, reconstruction quality decreases more smoothly as the noise increases.
    
    As a use case for calibration, we applied the VMF mechanism to a range of NLP experiments and it achieved competitive results compared to the standard DP-SGD. In some cases, there were clear advantages for VMF for short texts.

    Our work has raised several questions.  While the reconstruction attacks we examined show useful behaviour for calibration, they are narrower than MIAs in their applicability, and it is still an open question how well ROUGE-L scores relate to human judgements about reconstruction.  Given the increasing use of DP within NLP, and questions about actual privacy leakage suffered, we hope that this will prompt future work in this area.


\section*{Acknowledgment}
This project was undertaken with the assistance of resources and services from the National Computational Infrastructure (NCI), which is supported by the Australian Government. This work was also funded by the International Macquarie University Research Excellence Scholarship (iMQRES).

\appendix

\section{}

\subsection{Proofs}

\subsubsection{Proof of Corollary 1}\label{proof:1}

 \begin{proof}
The standard properties of differential privacy \cite{dwork-roth:2014}  result in the following relationship:        
    \[
     \VM^*(\epsilon, v)(Y) ~\leq e^{\epsilon \sum_{1\leq i\leq i'}d_2(\lambda u_i, \lambda'u_i')} \VM(\epsilon, v')(Y)~.
    \]
Observe that for any orthonormal set of vectors $u_i$ on the unit sphere, we have that  $\sum_{0\leq i \leq m}d_2(0, \lambda_iu_i)\leq \sum_{0\leq i \leq m} |\lambda_i| \sqrt m$. The result now follows using the triangle inequality of $d_2$, and that $d_2(\lambda_i u_i, \lambda'_iu_i')\leq 2$.     
    \end{proof}

\subsubsection{Proof of Theorem 2}\label{proof:2}

\begin{proof}
Line 10 of \Alg{alg:sgd2} ensures each vector in $B, B'$ lies on the unit sphere (since $C = 1$) and line 13 applies VMF parametrised by $\sigma$.
Applying \Thm{thm:d2_priv} to every vector in $B, B'$ yields \Eqn{e1720} since the (parallel) composition of $d_2$-privacy mechanisms gives a guarantee
with $Pr({\mathcal VM}(B) \in Z) \leq Pr({\mathcal VM}(B') \in Z) \times e^{\sigma \sum_{1\leq i\leq n}d_2(v_i, v_i')}$; this reduces to \Eqn{e1720} since all but one of the distances $d_2(v_i, v'_i)=0$, by assumption.
 The averaging in line 13 is an example of a  post-processing step which, by the data processing inequality property of $d$-privacy does not decrease the privacy guarantee \cite{fernandes:22:CSF}.

%
%
%
\end{proof}

\subsubsection{Proof of Corollary 2}\label{proof:3}

\begin{proof}
Observe that $\max_{B \sim B'} d_2(v, v') = 2$ since $v, v'$ lie on the unit sphere. $\epsilon$-DP on batches follows from choosing $\sigma=\frac{\epsilon}{2}$ which is the standard DP-SGD tuning from \cite{song-etal:2013}. Since batches are disjoint, the result follows by parallel composition for adjacent training sets $D, D'$. 
\end{proof}

\subsubsection{Proof of Corollary 3}\label{proof:4}

\begin{proof}
Follows from the fact that $d_2$-privacy implies $d_\theta$-privacy (since $d_2 \leq d_\theta$ pointwise on the unit sphere), using $d_\theta$ reasoning in \Thm{l1647} and using the same $\sigma$ tuning as per Cor~\ref{cor:epsdp}.
\end{proof}

\subsection{Hyperparameters}\label{app:environment_details}
   

   Table \ref{tab:hyperparameters} shows the hyperparameters used in the experiments. All experiments used the AdamW optimiser. If something is omitted, Pytorch's or Transformers' defaults were used. 

    \begin{table}[h!]
        \centering
        \small
        \setlength\tabcolsep{3.5pt} 
        \begin{tabular}{cccccc}
        \hline
        \textbf{Model} & 
        \textbf{Dataset} &
        \textbf{Batch} &
        \textbf{LR} &
        \textbf{Epochs} &
        \textbf{Seq len} \\ \hline
         \multirow{3}{*}{GPT2} & CoLA & 128  & 3e-4  & 30   & 40 \\
          & IMDb & 128  & 3e-3  & 5   & 256 \\
          & SST2 &  128  &  2e-3  &  3  & 66 \\\cline{2-6}
         \multirow{3}{*}{BERT} & CoLA & 128  & 5e-6  &  10  & 40 \\
          & IMDb & 128  &  3e-3 & 5  & 256  \\
          & SST2 &  128  &  2e-5  &  4    &66\\
         \hline
        \end{tabular}
        \caption{\textmd{Hyperparameters used.}}
        \label{tab:hyperparameters}
    \end{table}
    
\subsection{Datasets splits}\label{app:datasets_splits}

    Table \ref{tab:dataset_splits} shows the splits for each dataset. We also show the splits used for the MIA-R attacks, which trains reference models, thus requiring more dataset splits. We avoid very small splits for test sets in CoLA and SST2 in MIA-R by combining their training and dev sets, splitting equally into train/test sets, then spreading across the 10 reference and 1 target models. 

    \setlength{\tabcolsep}{3pt}
    \begin{table}[h!]
        \centering
        \small
        \begin{tabular}{cccc|cc}
        \hline
        \textbf{Dataset} &
        \textbf{Train} &
        \textbf{Validation} &
        \textbf{Test} & \textbf{Train} &
        \textbf{Test}  \\ \hline
         CoLA &5,056 &324 &516 & 259  &259 \\
         IMDb & 22,500 & 2,500& 25,000& 2,045 & 2,045\\
         SST2 & 53,710 & 5,850 & 872 & 3,100  & 3,100\\
         \hline
        \end{tabular}
        \caption{\textmd{Sizes of dataset splits for each model (11 total) in the MIA-R experiments (right), and all other experiments (left).}}
        \label{tab:dataset_splits}
    \end{table}

\subsection{Privacy budgets}\label{app:epsilons}

    For the utility experiments in Section \ref{sec:util_results}, we fixed the target $\epsilon$ for the utility experiments, and report the noise multiplier $\sigma$ returned by the Renyi privacy accountant \cite{8049725}, calculated based on the hyperparameters reported in Table \ref{tab:hyperparameters} and training sizes of Table \ref{tab:dataset_splits}. Table \ref{tab:budgets} shows the mapping to $\epsilon$ to the noise multipliers $\sigma$ for each dataset.

    \begin{table}[h!]
        \centering
        \small
        \begin{tabular}{ccccccccc}
        \hline
        \multirow{3}{*}{\textbf{Model}} &
        \multirow{3}{*}{\textbf{Dataset}} &
        \multicolumn{7}{c}{\textbf{Target} $\epsilon$}  \\
        &&$10^0$&$10^1$&$10^2$&$10^3$&$10^4$&$10^5$&$10^6$\\ \cline{3-9}
        &&\multicolumn{7}{c}{\textbf{Noise multiplier} $\sigma$}  \\\hline

        \multirow{3}{*}{\textbf{GPT2}} &CoLA &3.06&.747&.347&.173&.112&.066&.024\\
        &IMDb &1.01&.472&.226&.121&.089&.052&.019\\
        &SST2&.957&.458&.234&.127&.096&.072&.034\\
        \cline{2-9}

        \multirow{3}{*}{\textbf{BERT}} &CoLA&1.191&.615&.278&.140&.092&.043&.014\\
        &IMDb&.993&.460&.216&.116&.086&.072&.016\\
        &SST2&1.00&.478&.254&.137&.101&.078&.042\\
        \hline
    \end{tabular}
        \caption{\textmd{Mapping from $\epsilon$ to reported noise multipliers $\sigma$ for the Gaussian mechanism.}} 
        \label{tab:budgets}
    \end{table}

\subsection{Supplementary results}
\label{sec:supp-material}

\begin{figure*}[h!]
    \centering
    \begin{tabular}{ccc}
               
        \subfloat[MCC for GPT2 on CoLA]{\includegraphics[width=0.3\textwidth]{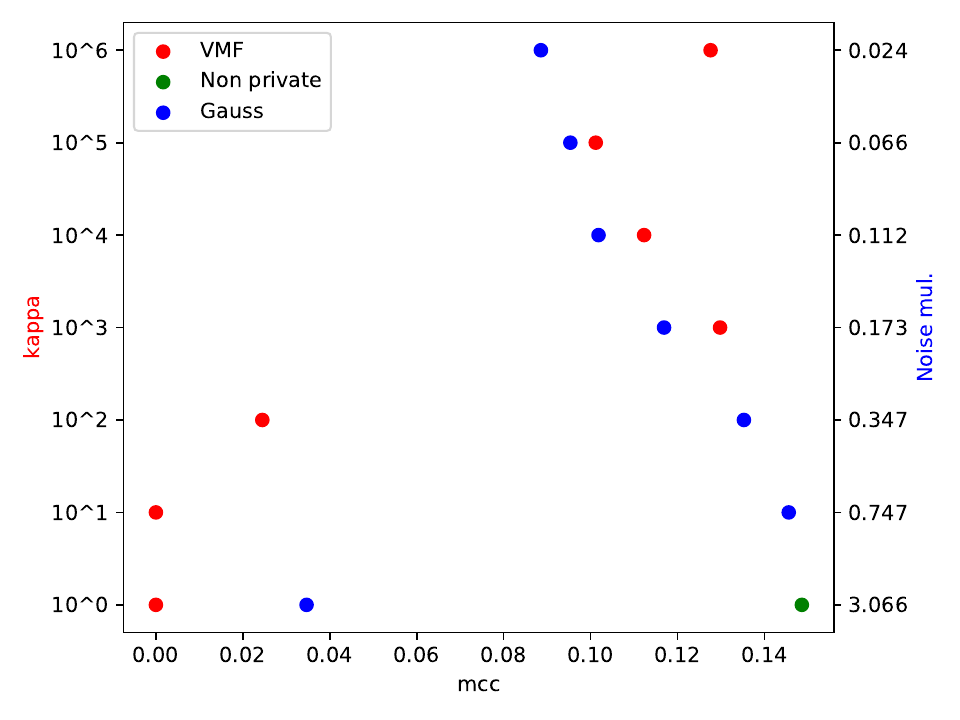}} &
        
        \subfloat[Accuracy for GPT2 on IMDb]{\includegraphics[width=0.3\textwidth]{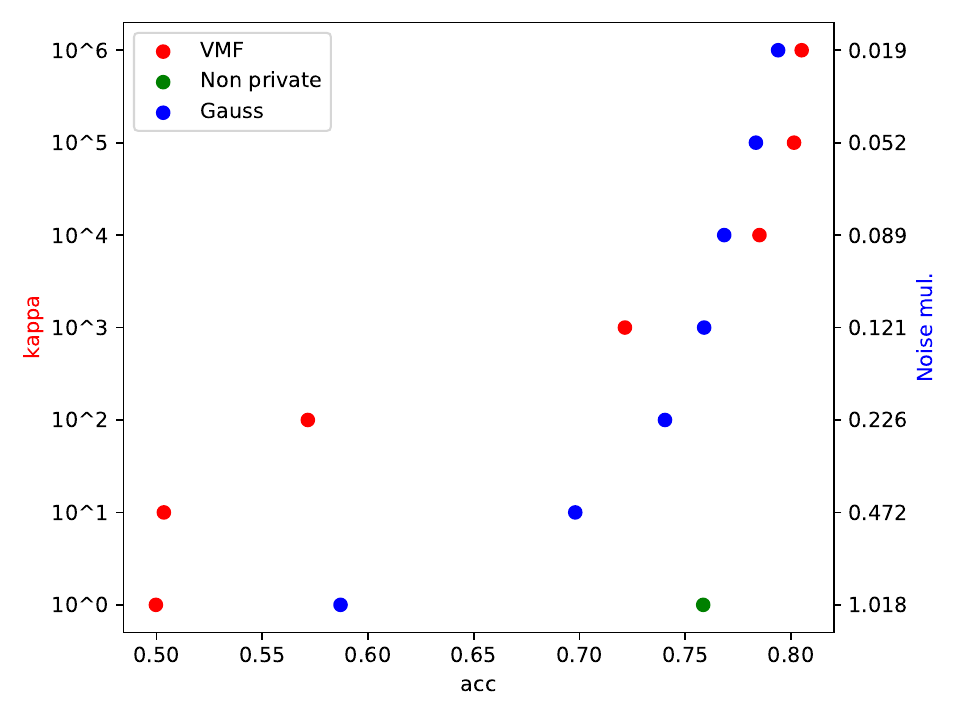}} &
    
        \subfloat[Accuracy for GPT2 on SST]{\includegraphics[width=0.3\textwidth]{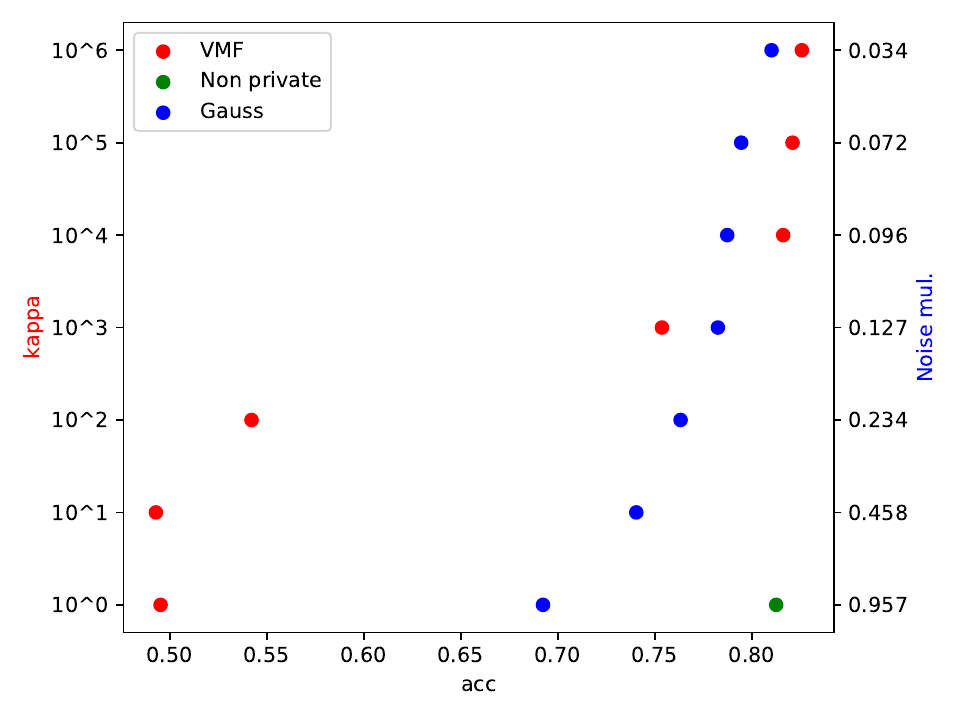}}  \\

        \subfloat[MCC for BERT on CoLA]{\includegraphics[width=0.3\textwidth]{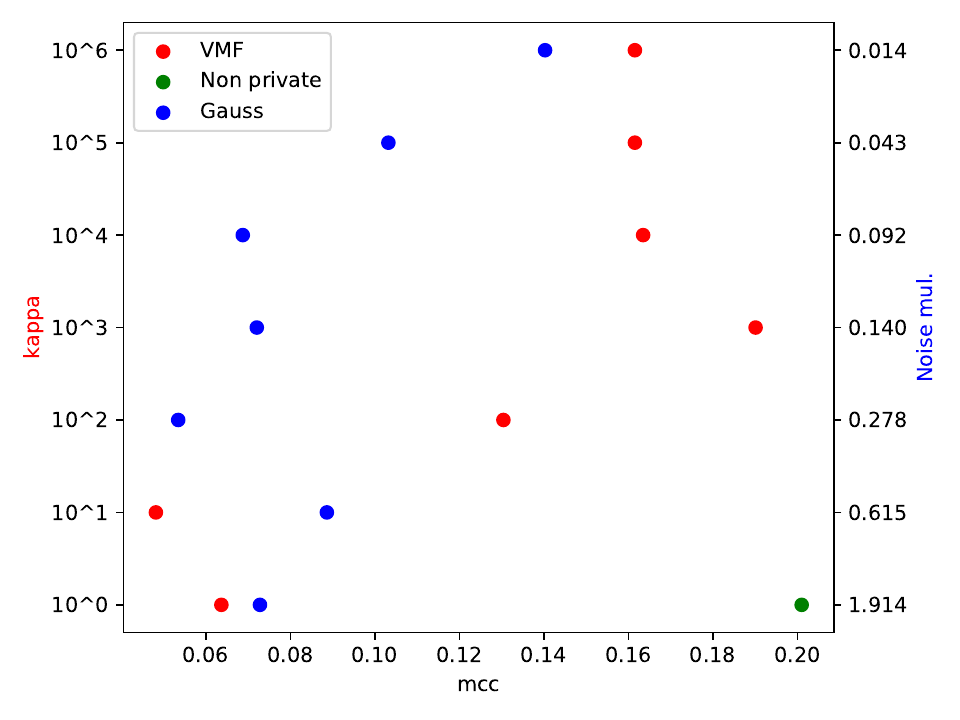}} &
        
        \subfloat[Accuracy for BERT on IMDb]{\includegraphics[width=0.3\textwidth]{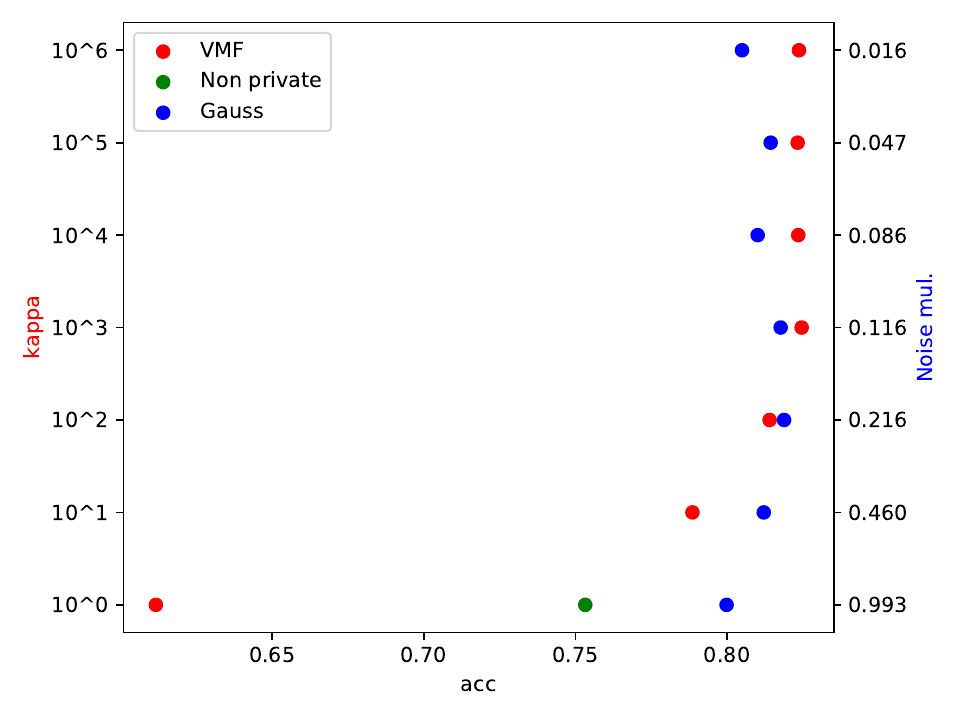}} &
        
        \subfloat[Accuracy for BERT on SST]{\includegraphics[width=0.3\textwidth]{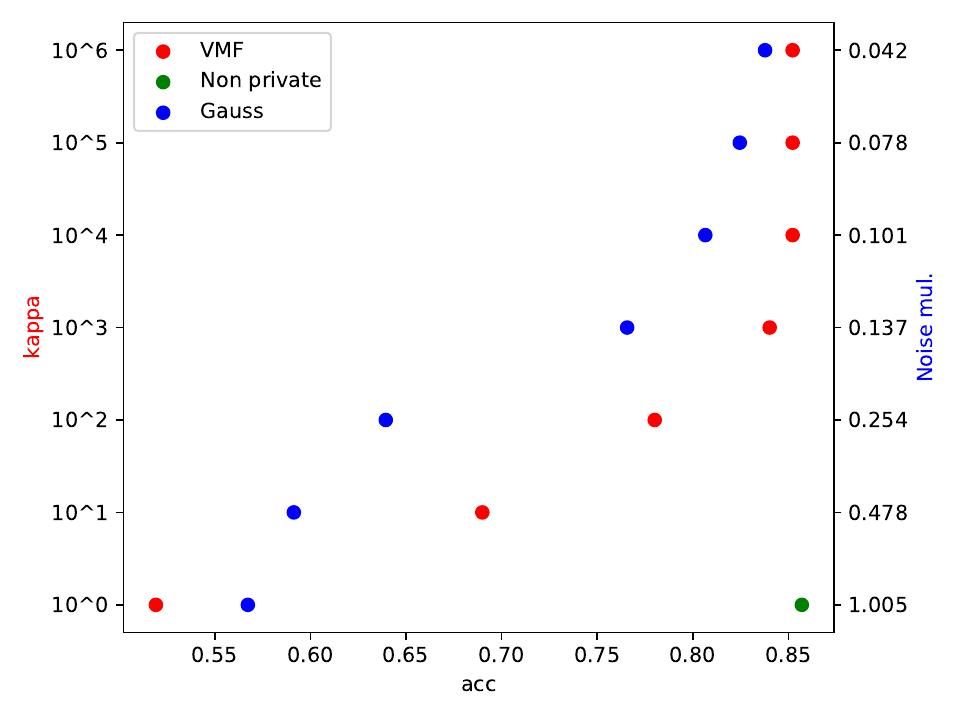}} 
    \end{tabular}
    \caption{\textmd{Utility results, in terms of accuracy or MCC, across the models and datasets under different privacy settings.}}\label{fig:utility_plots}
    \end{figure*}

    Fig~\ref{fig:utility_plots}) shows that utility tends to increase as noise diminishes (which for VMF is as $\kappa$ increases), with an occasional erratic behaviour as noted for Gaussian.

    Table~\ref{tab:lamp_examples_rouge} gives a few examples of reconstructed sentences that have the same ROUGE-L scores. Even when ROUGE-L scores are similar, humans might judge reconstructions to be of differing quality, in part because ROUGE-L does not distinguish among parts-of-speech (so reconstructing a determiner is as good as reconstructing a key noun). Additionally, reconstructions often contain non-words that give strong clues about the original but that ROUGE-L with its token matching does not pick up (e.g. ``who has seen my snorkel?'' reconstructed with noise as ``whokel scene has mynorkel?'').

    \begin{table}[h!]
        \centering
        \small
        \begin{tabular}{ccp{0.65\linewidth}}
        \hline
        \textbf{ROUGE-L} &
        \textbf{Privacy} &
        \textbf{Text} \\\hline
        \multirow{5}{*}{0.222} &
           \makecell{Original \\ CoLA}  & who has seen my snorkel?\\
           & $\sigma$ = .014 & whokel scene has mynorkel?\\
           & \makecell{Original \\ CoLA}  & whether she will win is a question mary never considered.\\
           & $\kappa$ = $10^5$ & mary. mary win whether although whether question. has ; \\

          \hline
                  \multirow{5}{*}{0.167} &

          \makecell{Original \\ SST2} & pummel us with phony imagery or music  \\
          &  $\sigma$ = .042 & imagery with musicenciesmmel usic orony  \\
          & \makecell{Original \\ SST2} & of softheaded metaphysical claptrap 
  \\
          & $\kappa$ = $10^2$ & of sells overlooking stem clamped uefa consonants curran \\

          \hline
                  \multirow{7}{*}{0.333} &

          \makecell{Original \\ SST2} & rotting underbelly 
\\
          & $\sigma$ = .042 &  [CLS]belly rotting rotting 
\\
          & \makecell{Original \\ SST2} & all the queen's men is a throwback war movie that fails on so many levels, it should pay reparations to viewers. 
\\
          & $\kappa$ = $10^5$ & onions tears all men is equal. abacks viewers men should'' a war movie often cut fails queenion long viewersions movie. 
\\
           \hline
         \end{tabular}
        \caption{\textmd{Reconstructed texts by LAMP against BERT, paired by same ROUGE-L score. $\kappa$ and $\sigma$ refer to VMF and Gaussian noises respectively. All gradients were shared.}} 
        \label{tab:lamp_examples_rouge}
    \end{table}

\bibliographystyle{unsrt}
\bibliography{custom}

\begin{IEEEbiography}
[{\includegraphics[width=1in,height=1.25in,clip,keepaspectratio]{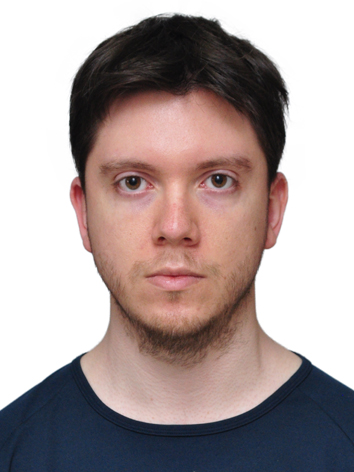}}]{Pedro Faustini} is a postdoctoral research associate at the School of Computing, Macquarie University. Previously, he completed his PhD also at Macquarie University, his Master's at the Federal University of ABC (UFABC) and bachelor's at the Federal University of Rio Grande do Sul (UFRGS). His past publications were accepted in prestigious venues in NLP, such as EMNLP, ACL and Interspeech. His research interests cover the intersections between privacy and NLP.  

\end{IEEEbiography}

\begin{IEEEbiography}
[{\includegraphics[width=1in,height=1.25in,clip,keepaspectratio]{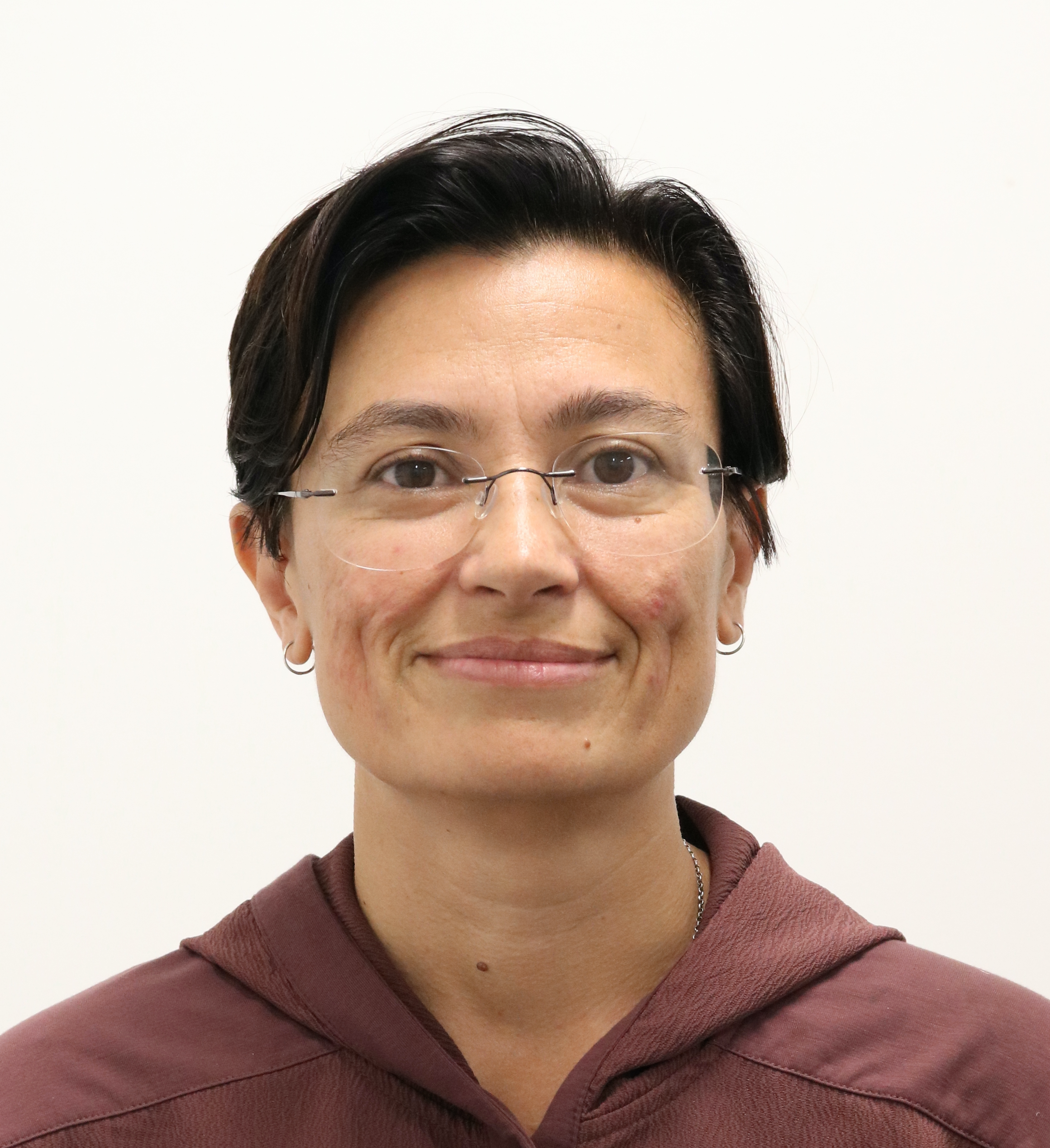}}]{Natasha Fernandes} is a senior lecturer in the School of Computing at Macquarie University, Australia.  Her research areas include differential privacy, metric differential privacy, quantitative information flow for privacy and privacy-preserving natural language processing.  She won the John Makepeace Bennett Award for Best Computing PhD Dissertation in Australia \& NZ for her doctoral research on these topics.
\end{IEEEbiography}

\begin{IEEEbiography}
[{\includegraphics[width=1in,height=1.25in,clip,keepaspectratio]{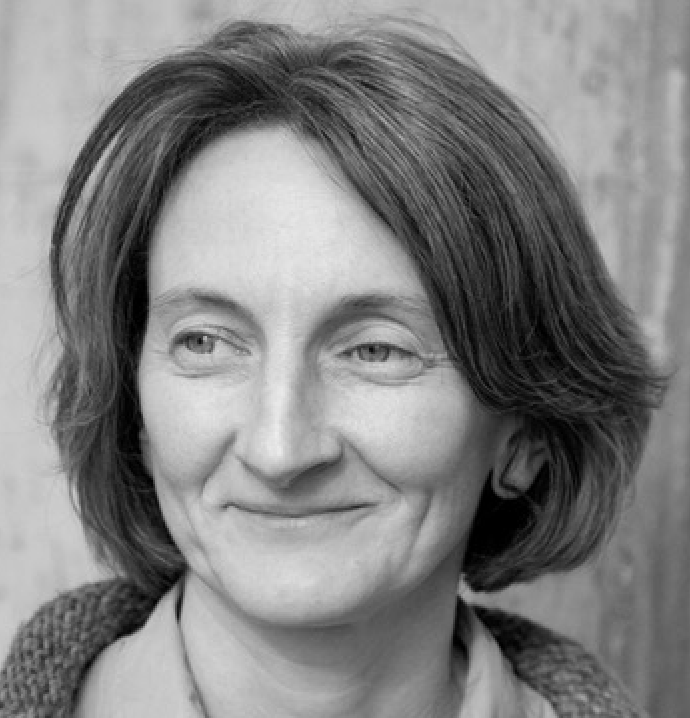}}]{Annabelle McIver} is a professor of Computer Science at Macquarie University in Sydney, and co-director of the Future Communications Research Centre. Annabelle trained as a mathematician at Cambridge and Oxford Universities. Her research uses mathematics to prove quantitative properties of programs, and more recently to provide foundations for quantitative information flow for analysing security properties. She is co-author of the book "Abstraction, Refinement and Proof for Probabilistic Systems", and "The Science of Quantitative Information Flow".
\end{IEEEbiography}

\begin{IEEEbiography}
[{\includegraphics[width=1in,height=1.25in,clip,keepaspectratio]{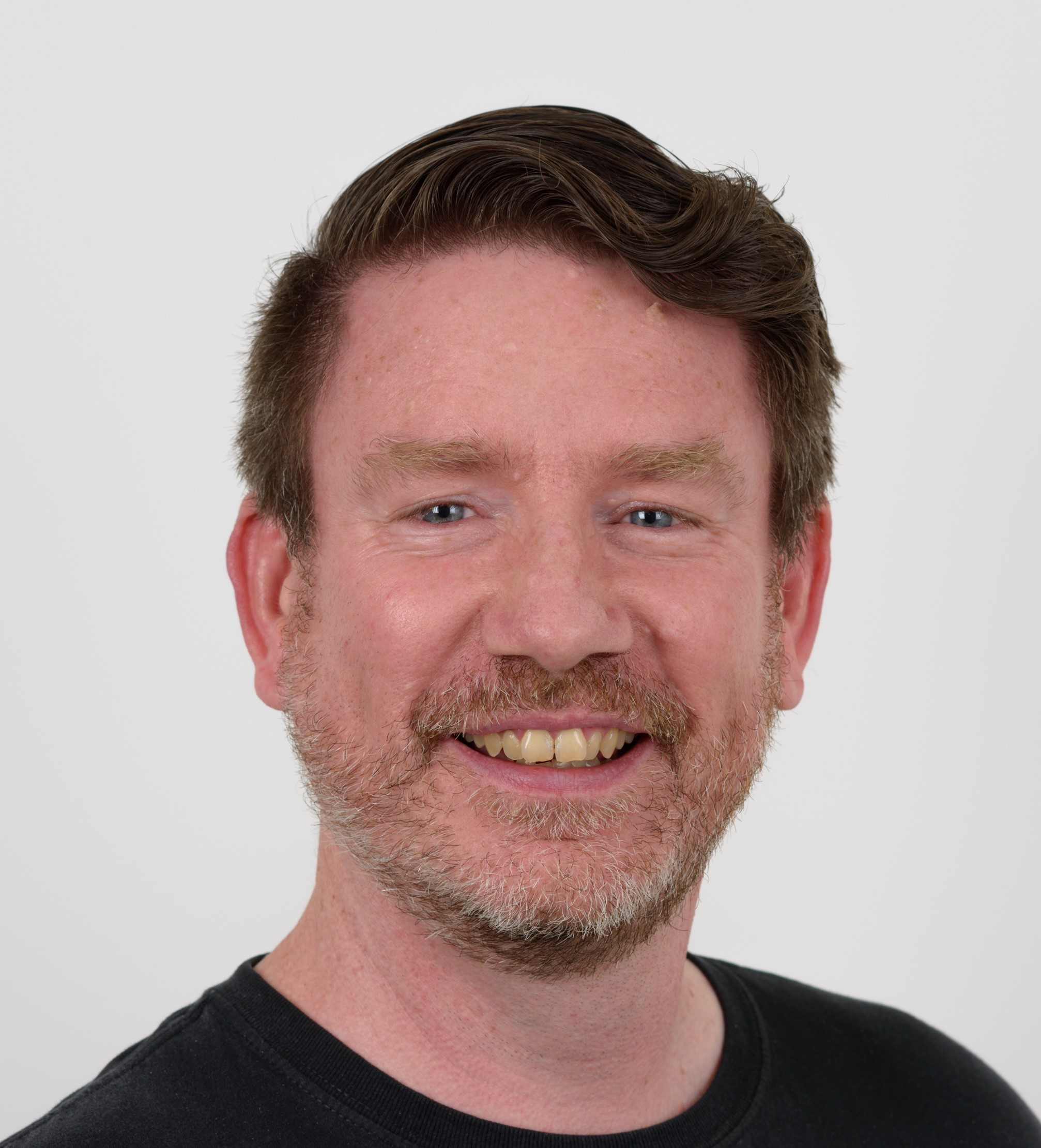}}]{Mark Dras} is a professor in the School of Computing at Macquarie University, Australia.  He works on machine learning and artificial intelligence, with a particular focus on Natural Language Processing and Large Language Models, especially on topics of privacy and security.  Mark is currently Treasurer of the Asia-Pacific Chapter of the Association for Computational Linguistics, the premier international scientific and professional society for people working on computational treatment of human language. He has also been appointed to the Australian Research Council’s College of Experts (2025-2027).

\end{IEEEbiography}






\EOD

\end{document}